\documentclass[a4paper,fleqn]{cas-dc}

\makeatletter
\def\@shortauthors{}
\def\@shorttitle{}
\makeatother

\usepackage{multirow}
\allowdisplaybreaks[4]
\usepackage{fancyhdr}
\usepackage{float}
\usepackage{xhfill}
\usepackage{graphics}
\usepackage{subcaption}
\usepackage{amssymb}
\usepackage{color}
\usepackage{bbding}

\usepackage{amsmath}
\usepackage{amssymb}
\usepackage{amsthm}
\usepackage{placeins}
\usepackage{booktabs}
\usepackage{colortbl}
\usepackage{xcolor}
\usepackage{caption}
\usepackage{graphicx}
\usepackage{url}
\usepackage{threeparttable}
\usepackage{microtype}
\DeclareUnicodeCharacter{211D}{\mathcal{L}}
\usepackage{flushend}

\newcommand{\blue}[1]{#1}

\makeatletter
\NewDocumentEnvironment{xyzenv}{ +b }{%
  \ifabc
    \expandafter\@firstofone
  \else
    \expandafter\@gobble
  \fi
  {#1}%
}{}
\renewcommand{\fnum@figure}{Fig. \thefigure.\@gobble}
\makeatother

\hypersetup{colorlinks=true, linkcolor=cyan, citecolor=cyan}

\usepackage[numbers,sort&compress]{natbib}

\usepackage{stfloats}
\usepackage{cleveref}

\renewcommand{\eqref}[1]{Eq. (\ref{#1})}
\addcontentsline{toc}{section}{Title of the section}

\usepackage[utf8]{inputenc}
\usepackage{tikz}
\usepackage{verbatim}
\usetikzlibrary{matrix, arrows.meta}
\usetikzlibrary{shapes.gates.logic.US, trees,positioning,arrows}

\tikzset{
  centered/.style = { align=center, anchor=center },
     empty/.style = { font=\sffamily\Large, centered, text width=2cm },
       box/.style = { font=\sffamily, fill=green, centered },
    result/.style = { font=\sffamily\scriptsize, fill=black!20, centered},
     arrow/.style = { very thick,->, >=Triangle},
}

\tikzset{global scale/.style={
    scale=#1,
    every node/.append style={scale=#1}
  }
}

\tikzset{
  dot/.style={
    circle, fill=black, inner sep=1pt, outer sep=0pt
  },
  dot label/.style={
    circle, inner sep=0pt, outer sep=1pt
  },
  pics/right angle/.append style={
    /tikz/draw, /tikz/angle radius=5pt
  }
}
\ExplSyntaxOn
\cs_gset:Npn \printorcid {}
\ExplSyntaxOff

\ExplSyntaxOn
\cs_gset:Npn \__first_footerline: {}
\ExplSyntaxOff

\begin{document}
\pagestyle{plain}

\fancyhead[L]{Fan, Jiang, Liu, Xue, Zhang and Liu}
\fancyhead[R]{}

\title [mode = title]{UltraLBM-UNet: Ultralight Bidirectional Mamba-based Model for Skin Lesion Segmentation}

\author[1]{Linxuan Fan}[type=editor,]
\fnmark[1]
\credit{Methodology, Investigation, Coding / Programming, Validation, Visualization, Writing-original draft}
\ead{linxuan.fan@vanderbilt.edu}

\author[2]{Juntao Jiang}[type=editor,]
\fnmark[1]
\credit{Conceptualization, Methodology, Investigation, Visualization, Writing-original draft, Project administration, Funding acquisition}
\ead{juntaojiang@zju.edu.cn}

\author[3]{Weixuan Liu}
\credit{Investigation, Visualization, Writing–review \& editing}
\ead{wxliu@stu.ecnu.edu.cn}

\author[2]{Zhucun Xue}
\credit{Investigation, Writing-review \& editing}
\ead{12432038@zju.edu.cn}

\author[2]{Jiajun Lv}
\credit{Investigation, Writing-review \& editing, Funding acquisition}
\ead{lvjiajun314@zju.edu.cn}

\author[2]{Jiangning Zhang}
\credit{Methodology, Writing-review \& editing, Supervision}
\ead{186368@zju.edu.cn}

\author[2]{Yong Liu}
\cormark[1]
\credit{Conceptualization, Project administration, Supervision}
\ead{yongliu@iipc.zju.edu.cn}

\address[1]{Data Science Institute, Vanderbilt University, Nashville, TN 37240, USA}

\address[2]{College of Control Science and Engineering, Zhejiang University, Hangzhou, Zhejiang 310000, China}

\address[3]{School of Computer Science and Technology, East China Normal University, Shanghai, 200062, China}
\fntext[1]{These authors contributed equally to this work.}
\cortext[cor1]{Corresponding author}

\begin{abstract}
Skin lesion segmentation is a crucial step in dermatology for guiding clinical decision-making. However, existing methods for accurate, robust, and resource-efficient lesion analysis have limitations, including low performance and high computational complexity. To address these limitations, we propose UltraLBM-UNet, a lightweight U-Net variant that integrates a bidirectional Mamba-based global modeling mechanism with multi-branch local feature perception. The proposed architecture integrates efficient local feature injection with bidirectional state-space modeling, enabling richer contextual interaction across spatial dimensions while maintaining computational compactness suitable for point-of-care deployment. Extensive experiments on the ISIC 2017, ISIC 2018, and PH\textsuperscript{2} datasets demonstrate that our model consistently achieves state-of-the-art segmentation accuracy, outperforming existing lightweight and Mamba counterparts with only 0.034M parameters and 0.060 GFLOPs. In addition, we introduce a hybrid knowledge distillation strategy to train an ultra-compact student model, where the distilled variant UltraLBM-UNet-T, with only 0.011M parameters and 0.019 GFLOPs, achieves competitive segmentation performance.
These results highlight the suitability of UltraLBM-UNet for point-of-care deployment, where accurate and robust lesion analyses are essential. \blue{Furthermore, on-device inference experiments on a consumer smartphone demonstrate that UltraLBM-UNet-T achieves 18.2\,ms latency (54.9\,FPS) and UltraLBM-UNet achieves 20.8\,ms (48.1\,FPS) at $256\times256$ resolution with batch size 1 via ONNX Runtime, ranking first and second respectively among all evaluated methods in mobile deployment speed.} The source code is publicly available at \url{https://github.com/LinLinLin-X/UltraLBM-UNet}.

\end{abstract}

\begin{highlights}
\item Propose a lightweight Global–Local Multi-branch Perception Module that effectively integrates global contextual dependencies and fine-grained local representations.

\item Demonstrate that a bi-directional Mamba with shared weights can enhance information flow without increasing parameters.

\item Introduce a hybrid knowledge distillation strategy to compress the model further and improve the performance of ultra-compact variants without additional inference cost.

\item Achieve state-of-the-art segmentation performance on ISIC 2017, ISIC 2018, and PH\textsuperscript{2} datasets with the smallest parameter count and computational cost.

\item \blue{Validate real-world deployability through on-device inference experiments on a consumer smartphone, where both UltraLBM-UNet and UltraLBM-UNet-T achieve the fastest inference speed among all compared methods via ONNX Runtime.}
\end{highlights}


\setlength{\parindent}{2em}
\setlength{\mathindent}{0em}

\tnotemark[1]
\tnotetext[1]{
This is the accepted manuscript of an article published in Neurocomputing.
The final published version is available at
\url{https://doi.org/10.1016/j.neucom.2026.134608}.
\copyright~2026.
This manuscript version is made available under the
CC BY-NC-ND 4.0 license
(\url{https://creativecommons.org/licenses/by-nc-nd/4.0/}).
}

\maketitle

\section{Introduction}

Skin cancer is one of the most common malignancies worldwide~\cite{siegel2024cancer}, and its most aggressive type, melanoma, accounts for the majority of skin-cancer-related deaths~\cite{gershenwald2018melanoma}. Early and accurate delineation of malignant regions is crucial for improving patient outcomes and guiding clinical management. Skin lesion segmentation aims to delineate diseased areas from surrounding healthy tissue, forming the basis for lesion quantification, classification, and monitoring. Traditional manual delineation is time-consuming and subjective, motivating the development of data-driven methods~\cite{mirikharaji2023survey}. The rise of deep learning has demonstrated dermatologist-level accuracy in skin lesion analysis~\cite{esteva2017dermatologist}, highlighting its potential for automated segmentation and improving computer-assisted diagnosis, thereby ultimately benefiting patient care.

The growing emphasis on point-of-care (POC) healthcare, which delivers medical services directly at the patient’s side, underscores the need for rapid and accessible diagnostic tools in diverse clinical settings~\cite{zaki2024role}. In dermatology, the integration of POC imaging, such as handheld dermatoscopes or mobile phone–based systems, enables real-time skin assessment and highlights the importance of automated lesion segmentation for timely clinical decision-making. Skin cancer diagnosis, in particular, benefits significantly from POC workflows, as early identification of malignant lesions can directly influence treatment options and patient prognosis. Unlike well-equipped hospital imaging centers, POC environments frequently face constraints such as limited computational capability, battery-dependent devices, unstable network access, and the need for instant response. These practical limitations make it challenging to deploy heavy transformer- or CNN-based models, thereby amplifying the necessity for lightweight, fast, and memory-efficient segmentation architectures. Consequently, designing robust yet efficient models is crucial for ensuring accurate, resource-aware, and widely deployable skin lesion analysis at the point of care.
\begin{figure*}[h]
  \centering
  \includegraphics[width=0.9\linewidth]{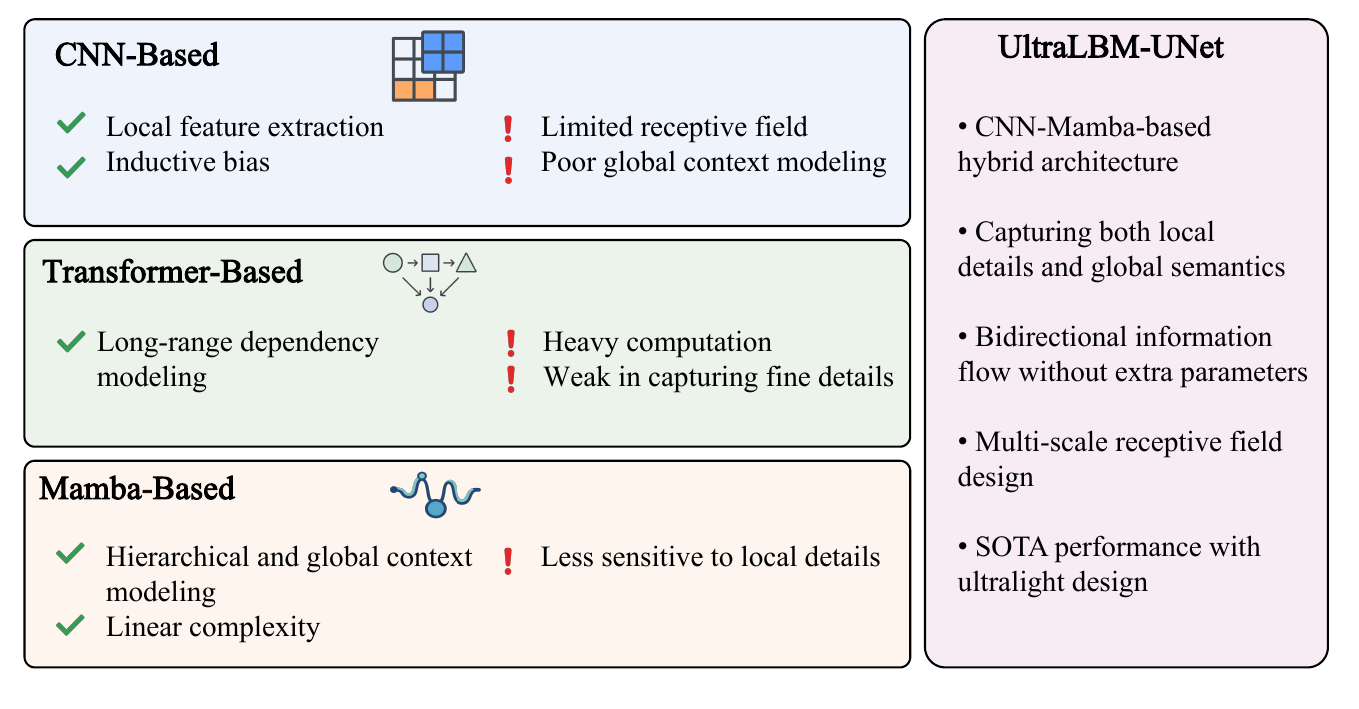}
  \caption{Comparative analysis of CNN-, Transformer- and Mamba-based models, highlighting their respective strengths and weaknesses, as well as the superiority of our UltraLBM-UNet.}
  \label{fig:combined_poc_teaser}
\end{figure*}

In the past decade, advances in medical image segmentation have been driven by U-Net~\cite{ronneberger2015u} and its variants ~\cite{milletari2016v,oktay2018attention,zhou2018unet++,huang2020unet,xiao2018weighted}. The encoder-decoder structure of U-Net, enhanced by skip connections, has proven highly effective in preserving fine spatial details within complex anatomical structures. Thanks to the success of transformer applications in vision tasks ~\cite{dosovitskiy2020image}, models like TransUNet~\cite{chen2021transunet}, \blue{MT-UNet~\cite{wang2022mixed}}, and Swin-Unet~\cite{cao2022swin}, which integrate transformers to capture long-range dependencies, have achieved new levels of segmentation accuracy. However, these advances come at the cost of increased computational complexity, which limits their feasibility in resource-constrained real-time settings. Linear attention-based sequence modeling methods, such as Mamba~\cite{gu2023mamba} and RWKV~\cite{peng2023rwkv}, have recently gained considerable attention and been widely adopted in 
medical image segmentation~\cite{ruan2024vm, ma2024u,jiang2025rwkvunetimprovingunetlongrange}. These models offer efficient long-range dependency modeling with linear computational complexity, making them particularly suitable for lightweight architectures and real-time clinical applications.

For point-of-care medical image segmentation, recent studies~\cite{valanarasu2022unext,ruan2022malunet,ruan2023ege,hu2024leanet,li2024lite, jiang2024lv} have focused on developing efficient U-Net variants that balance accuracy and computational cost, enabling real-time on-device inference in clinical settings. Lightweight segmentation models based on Mamba~\cite{gu2023mamba} have also been explored, aiming to balance efficiency and contextual modeling~\cite{liao2024lightm,wu2024ultralight}. In summary, the main techniques for lightweight UNet design include architectural simplification, efficient attention or feature aggregation, multi-scale fusion, and enhancing feature representation through re-parameterization or dynamic modules, all while minimizing computational redundancy.  However, many architectures still struggle to effectively integrate global context and local detail, which is essential for precise lesion delineation in complex skin images. The pursuit of lightweight design remains an ongoing challenge, as real-world clinical applications demand models that are not only more efficient but also more capable.

To address these limitations, we propose UltraLBM-UNet, a lightweight U-Net variant that achieves effective fusion of local and global representations through a bidirectional Mamba mechanism. The proposed architecture integrates efficient local feature injection with bidirectional state-space modeling, enabling richer contextual interaction across spatial dimensions while maintaining computational compactness suitable for point-of-care deployment. Extensive experiments on the ISIC 2017~\cite{codella2018skin}, ISIC 2018~\cite{codella2019skin,tschandl2018ham10000} and PH\textsuperscript{2}~\cite{mendoncca2015ph2} skin lesion segmentation benchmarks demonstrate that UltraLBM-UNet achieves state-of-the-art performance, surpassing existing lightweight and Mamba-based counterparts in both accuracy and efficiency. The comparative analysis of CNN-, Transformer-, and Mamba-based models, as well as the superiority of our UltraLBM-UNet, is shown in Fig. ~\ref{fig:combined_poc_teaser}.
\begin{figure*}[h]
  \centering
  \includegraphics[width=1\textwidth]{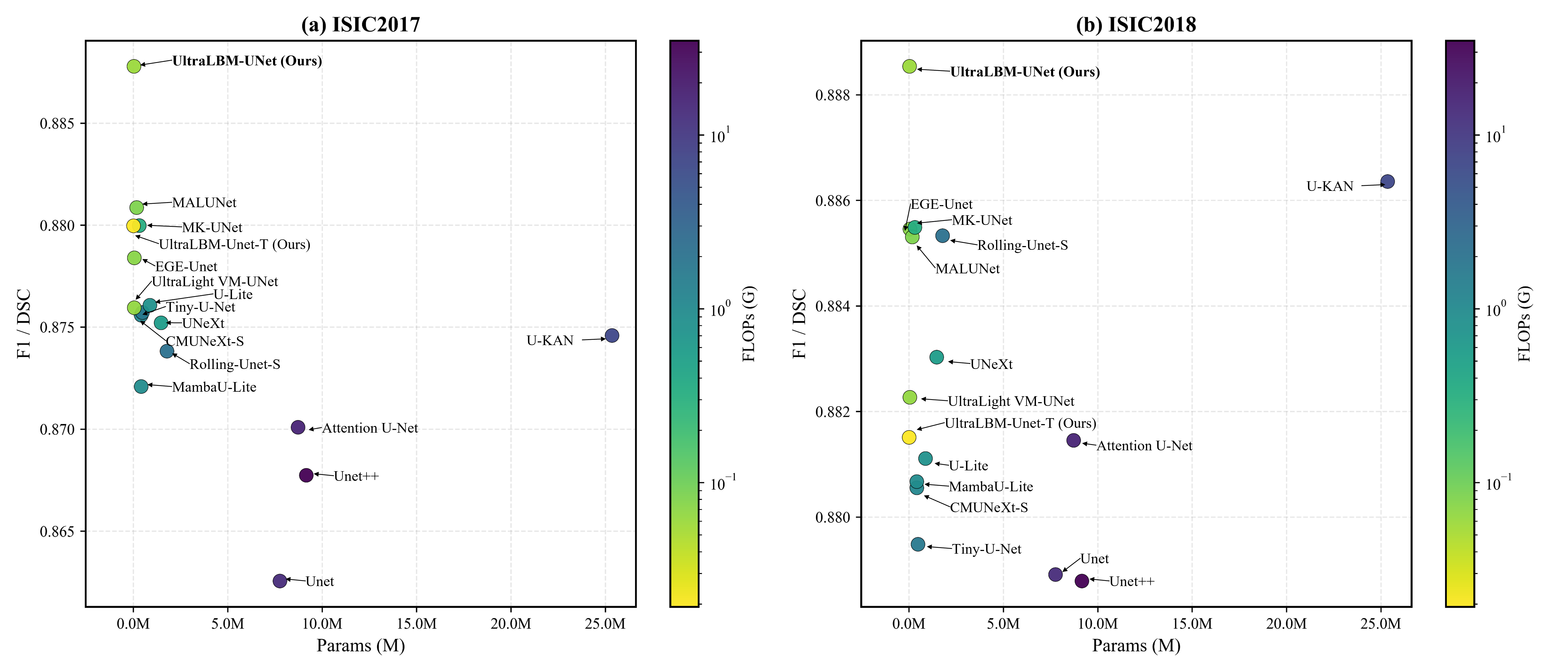}
  \caption{Comparison of segmentation models on ISIC2017 and ISIC2018. Each point represents a model, plotted by its parameter size (x-axis) and F1/DSC score (y-axis), with color indicating computational cost in GFLOPs.}
  \label{fig:param_vs_f1_dsc}
\end{figure*}
The main contributions of this work are summarized as follows:
\begin{itemize}
    \item We propose a lightweight \textbf{Global–Local Multi-branch Perception Module (GLMBP)}, which integrates global contextual dependencies and fine-grained local representations through a \blue{four-branch channel decomposition strategy with learnable scalar modulation, achieving adaptive and parameter-efficient feature routing beyond existing global-local designs.}
    
    \item We demonstrate that a \textbf{bi-directional Mamba with shared weights} can enhance information flow without increasing the number of parameters, achieving better global consistency and semantic completeness.

    \item We further introduce a \textbf{hybrid knowledge distillation strategy} to train an ultra-compact student model, where a carefully designed distillation loss transfers structural and boundary-aware knowledge from the full-capacity network without modifying the inference architecture.
    
    \item Extensive experiments on the \textbf{ISIC 2017}, \textbf{ISIC 2018} and \textbf{PH\textsuperscript{2}} datasets show that our model achieves the \textbf{state-of-the-art segmentation performance} with the smallest parameter count and computational cost among existing methods, as illustrated in Fig.~\ref{fig:param_vs_f1_dsc}.

    \item \blue{We validate the real-world deployability of UltraLBM-UNet by developing an Android application and conducting on-device inference experiments on a consumer smartphone via ONNX Runtime (Fig.~\ref{fig:app_screenshot}), demonstrating that both UltraLBM-UNet and UltraLBM-UNet-T achieve the fastest on-device inference speed among all compared methods.}
    
    \begin{figure*}[htp]
    \centering
    \includegraphics[width=0.45\linewidth]{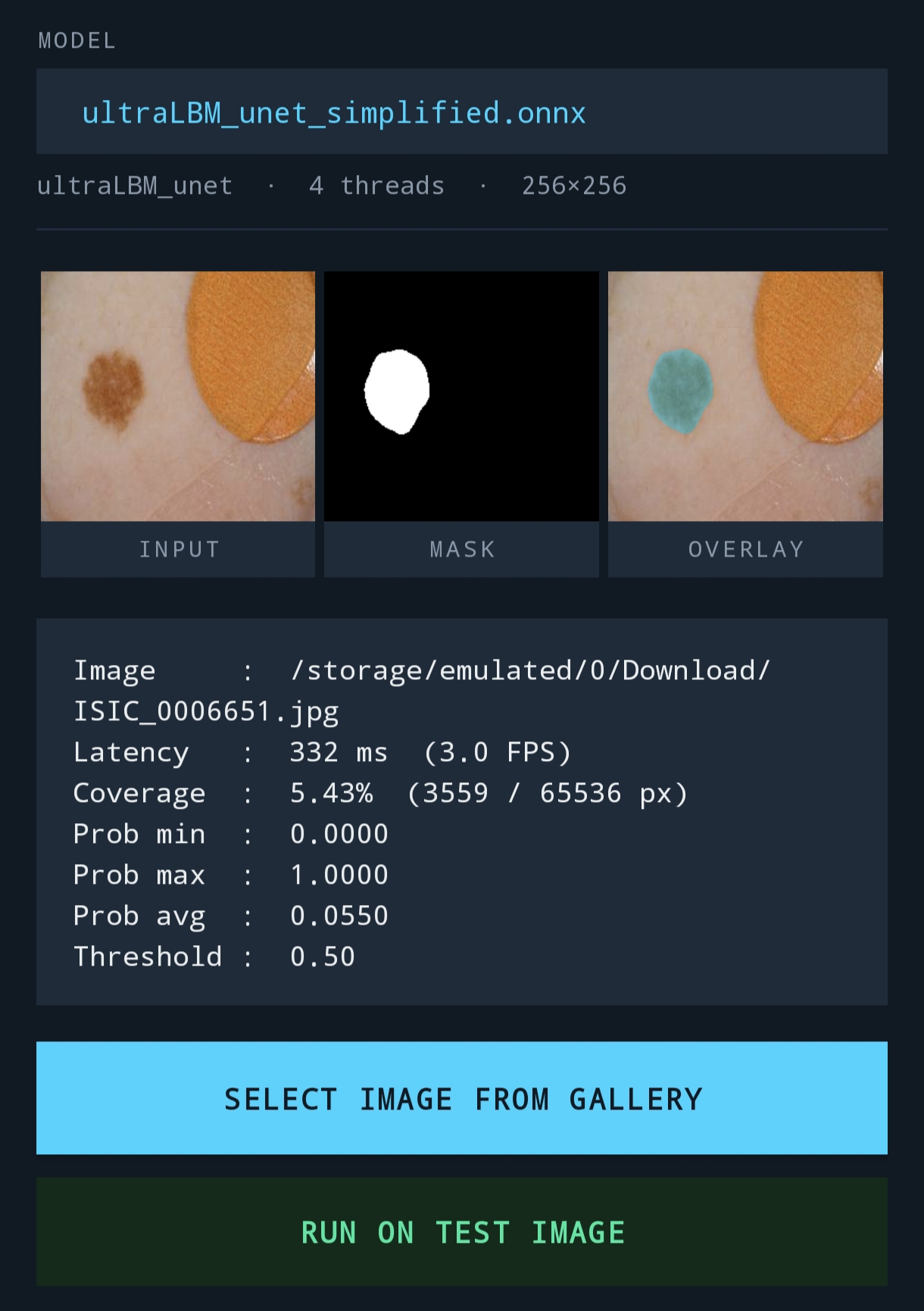}
    \caption{\blue{Screenshot of the developed Android application running UltraLBM-UNet on a consumer smartphone (Xiaomi 2311DRK48C, Android~14). The app visualizes the input dermoscopic image, binary segmentation mask, and cyan lesion overlay in real time via ONNX Runtime.}}
    \label{fig:app_screenshot}
\end{figure*}
\end{itemize}

\section{Related Work}

\subsection{Lightweight UNet Variants}

In the pursuit of efficient medical image segmentation, lightweight UNet variants have been developed to strike a balance between performance and computational demands. UNeXt~\cite{valanarasu2022unext} and LMU-Net~\cite{ma2024lmu} integrate MLPs to streamline the architecture, MAL-UNet~\cite{ruan2022malunet} employs multi-scale attention to capture features at various scales, EGE-UNet~\cite{ruan2023ege} leverages GHPA and GAB modules for diverse feature extraction and multi-scale fusion, Lite-UNet~\cite{li2024lite} combines gradient aggregation, graph correlation attention, and Ghost CBAM for rapid and precise localization, and LDSE-UNet~\cite{li2024dseunet} uses dynamic spatial group enhancement to achieve high accuracy with minimal parameters. LB-UNet~\cite{xu2024lb} combines an extremely compact GSA-based backbone with a novel boundary-enhanced PMA module that integrates region and boundary cues through auxiliary prediction and skip-connection fusion. LV-UNet~\cite{jiang2024lv} enhances feature representation through fusible blocks with spatially-aware nonlinear activations and achieves efficient inference via re-parameterization of convolutions and batch normalization. MK-UNet~\cite{rahman2025mk} uses a new multi-kernel depth-wise block and integrated attention to capture rich multi-scale features. A common design philosophy among these lightweight models is to reduce redundant computation while enhancing feature representation through efficient attention, aggregation, or fusion mechanisms. However, despite these innovations, most of these lightweight models still lack effective integration of global context, limiting their ability to capture long-range dependencies critical for precise lesion delineation.

\subsection{Mamba-based UNet Variants}

To address the local modeling limitation of CNNs and the quadratic complexity of Transformers, recent studies have turned to State Space Models (SSMs), particularly the Mamba architecture~\cite{gu2023mamba}. Mamba offers linear complexity while capturing long-range dependencies via a selective state space mechanism, making it well-suited for efficient vision tasks. Mamba-based designs such as VM-UNet~\cite{ruan2024vm}, Mamba-UNet~\cite{wang2024mamba}, and Swin-UMamba~\cite{liu2024swin} have demonstrated competitive accuracy. Moreover, lightweight variants reduce parameters and FLOPs by utilizing compact Vision Mamba layers. Specifically, LightM-UNet~\cite{liao2024lightm} compresses Mamba blocks, while UltraLight VM-UNet~\cite{wu2024ultralight} splits features into parallel branches for efficient long-range context modeling. However, existing models still face challenges:
\begin{itemize}
\item \textbf{Information Flow and Context:} The inherently unidirectional state transition in Mamba leads to asymmetric information propagation, where each position can only aggregate context from a single scanning direction. This limits the model’s ability to capture bidirectional spatial dependencies, resulting in incomplete context integration, particularly detrimental for tasks like segmentation that require precise boundary localization and global structural consistency.
\item \textbf{Global-Local Feature Integration:}  Many existing Mamba-based architectures emphasize global modeling while providing insufficient mechanisms to preserve or fuse fine-grained local details. This imbalance is especially problematic in medical image segmentation, where lesion edges and textural cues require strong local representational capability.
\end{itemize}

Therefore, in designing Mamba-based lightweight medical image segmentation models, key considerations include effectively combining local and global information, implementing multi-directional information flow, and further reducing both parameters and computational cost.

\section{Methodology: UltraLBM-UNet}

\begin{figure*}[htp]
    \centering
    \includegraphics[width=1\textwidth]{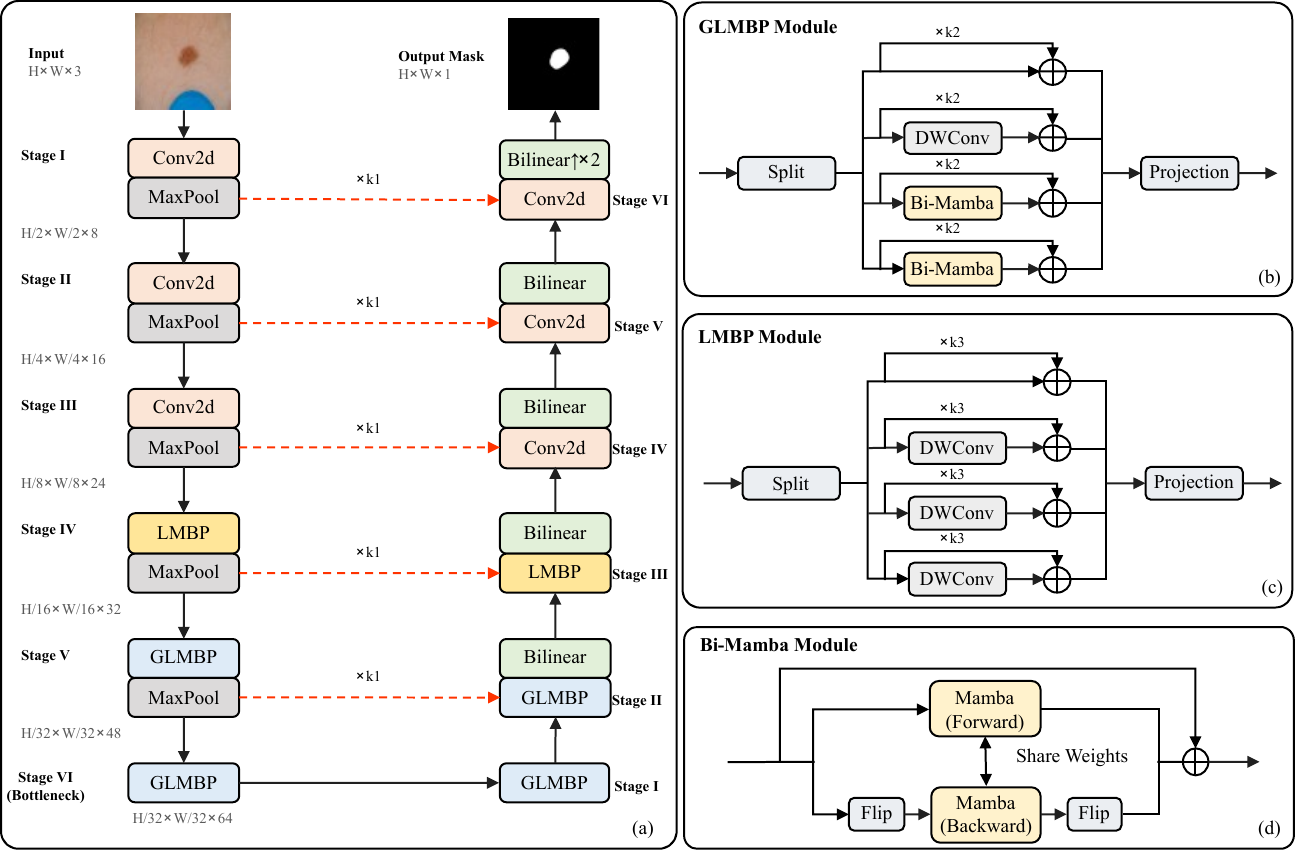} 
    \caption{The architecture of UltraLBM-UNet. (a) Overall UltraLBM-UNet Architecture (Encoder-Decoder) with skip-connections and layer dimensions. (b) LMBP Module used in Stage IV with parallel-branch decomposition and residual fusion of local features. (c) GLMBP Module, the core block, integrating Bi-Mamba (global context) and DWConv (local features) via parallel branches. (d) Bi-Mamba Module with shared weights, implemented via sequence flipping for non-causal bi-directional perception.}
    \label{fig:sample_png}
\end{figure*}

\subsection{Overall Architecture}

The overall architecture of UltraLBM-UNet can be seen in Fig.~\ref{fig:sample_png}. The designed model adopts an encoder-decoder architecture, comprising a 6-stage encoder, a decoder, and skip-connections. The channel dimensions are set to $[8, 16, 24, 32, 48, 64]$ across the layers. The encoder extracts features from the input image, reducing the spatial dimension to a minimum of $H/32 \times W/32$, while the decoder follows a symmetric setup, using the same modules and bilinear interpolation for upsampling to recover the resolution. The skip-connections are established between corresponding encoder and decoder stages, where feature maps are fused via element-wise addition to preserve spatial details. UltraLBM-UNet-T has the same architecture, \blue{with the channel dimensions of $[4, 8, 12, 16, 24, 32]$.} 

The encoder extracts features hierarchically: the first three layers use standard convolution blocks with max-pooling for shallow feature extraction and downsampling. Stages IV--VI employ the Global--Local Multi-branch Perception (GLMBP) module for feature extraction, combining Mamba for global context and depthwise separable convolutions for local details. Specifically, Stages IV and V also use max-pooling to downsample further. The decoder mirrors this structure, applying GLMBP in the first three upsampling stages and standard convolution blocks in the final two stages, with bilinear interpolation to gradually restore spatial resolution. 

\subsection{Global-Local Multi-Branch Perception Module}

The proposed GLMBP block is the core innovation of our model, designed through a unified modular architecture to collaboratively address two fundamental challenges in dense prediction tasks: capturing global context and preserving local details. This module integrates the global modeling capability of the SSM with the local perception advantage of CNN through a parallel multi-branch architecture, while strictly controlling the computational overhead. 

Let the input feature map be $X \in \mathbb{R}^{B \times C \times H \times W}$. It is first flattened into a sequence of shape $B \times N \times C$ ($N = H \cdot W$) and normalized along the channel dimension using Layer Normalization (LN), resulting in
$\tilde{X}_{\text{norm}}$. The core of the GLMBP module processes $\tilde{X}_{\text{norm}}$ in parallel. 
The sequence is evenly split along the channel dimension into four subsequences 
$X_1, X_2, X_3, X_4 \in \mathbb{R}^{B \times N \times C'}$, where $C' = C/4$:
\begin{equation}
\tilde{X}_{\text{norm}} = [X_1 \mid X_2 \mid X_3 \mid X_4].
\end{equation}
Here, $X_1$ and $X_2$ go to global modeling, $X_3$ handles local features, and $X_4$ acts as an identity branch.

\subsubsection{Bi-directional Mamba with Shared Weights}

To establish long-range dependencies among pixels across the entire image, we utilize the Mamba, the SSM of which offers linear complexity, making it highly efficient for processing long sequences resulting from flattening high-resolution images.
To construct a comprehensive, non-causal global context representation, we employ a Bi-directional Mamba mechanism for both $X_1$ and $X_2$. Both branches and their respective bidirectional paths share the same Mamba instance $M(\cdot)$. \blue{For $i \in \{1,2\}$, the backward path} is implemented via sequence flipping:
\begin{align}
M_{i, \text{fwd}} &= M(X_i), \\[3pt]
M_{i, \text{bwd}} &= \operatorname{Flip}\!\big(M(\operatorname{Flip}(X_i))\big).
\end{align}

The final branch output $X_i^{G}$  is formed by a residual connection scaled by a learnable scalar $\gamma \in \mathbb{R}$:
\begin{equation}
X_i^{G} = M_{i, \text{fwd}} + M_{i, \text{bwd}} + \gamma \cdot X_i.
\end{equation}

This scalar $\gamma$ allows the network to adaptively adjust the intensity of the fusion of global contextual information during training.

\subsubsection{Local Perception Branch}

While Mamba excels at global modeling, its sequence-based processing can diminish the perception of precise 2D spatial patterns. To compensate for this, the $X_3$ branch is specifically designed to enhance local feature representation. We employ the Depthwise Separable Convolution (DWConv) module to process $X_3$, which is reshaped back to its 2D format ($X_{3, \text{2d}} \in \mathbb{R}^{B \times C' \times H \times W}$). The output is fused with the original input via a residual connection as well:
\begin{equation}
X_3^{l} = \text{Reshape}(\text{DWConv}(X_{3, \text{2d}})) + \gamma \cdot X_3.
\end{equation}
\subsubsection{Identity Shortcut Branch}
The identity shortcut branch preserves the original feature information without any additional transformation. This not only ensures that fine-grained spatial details are retained throughout the GLMBP module, but also avoids introducing extra parameters or computational burden. Furthermore, the identity path provides an efficient route for gradient flow, stabilizing optimization and mitigating potential vanishing-gradient issues. By enabling direct feature reuse and eliminating redundant transformations, the identity shortcut effectively reduces both computational cost and parameter count while enhancing representational diversity. This design philosophy is consistent with lightweight architectures such as GhostNet~\cite{han2020ghostnet}, where efficient identity connections help maintain expressiveness with minimal overhead. Similar to the other branches, we include a learnable scaling factor $\gamma$ to control the contribution of this branch adaptively:
\begin{equation}
X_4^{I} = X_{4} + \gamma \cdot X_4.
\end{equation}

\subsubsection{Global-Local Information Fusion}

The outputs from the four branches, two global branches, one local branch, and one identity branch, are fused together via channel-wise concatenation.
\begin{equation}
\tilde{X}^{G}_{\text{fuse}} = [X_1^G \mid X_2^G \mid X_3^L \mid X_4^{I}],
\end{equation}
where $X_1^G$ and $X_2^G$ carry the bidirectional global contextual information, $X_3^L$ encodes multi-receptive-field local features, and $X_4^I$ preserves the original identity features.
\subsection{Local Multi-Branch Perception Module}

Local Multi-Branch Perception (LMBP) module retains the four-branch decomposition and residual fusion structure of GLMBP but replaces the bi-directional Mamba branches with DWConvs, which are configured with a fixed kernel size of 3, ensuring fine-grained local feature extraction without expanding the receptive field. All learnable branches operate purely in the local spatial domain, while the identity branch remains unchanged. LMBP is applied only in the shallow stage (Stage III), where local texture and boundary cues are most prominent, allowing the model to enhance fine-grained spatial details before deeper layers aggregate global semantic context. The feature map $X$ is evenly split along the channel dimension into four submaps after normalization:
$X_1, X_2, X_3, X_4 \in \mathbb{R}^{B \times C' \times H \times W}$, where $C' = C/4$:
\begin{equation}
\tilde{X}_{\text{norm}} = [X_1 \mid X_2 \mid X_3 \mid X_4].
\end{equation}
Here, $X_1$, $X_2$, and $X_3$ handle local features, and $X_4$ acts as an identity branch.

\begin{equation}
X_i^{L} = \text{Reshape}(\text{DWConv}(X_{i,\text{2d}})) + \gamma \cdot X_i,\quad i = 1,2,3
\end{equation}
The outputs from the four branches, three local branches, and one identity branch, are fused via channel-wise concatenation.
\begin{equation}
\tilde{X}^{L}_{\text{fuse}} = [X_1^L \mid X_2^L \mid X_3^L \mid X_4^{I}],
\end{equation}
where $X_4^{I} = X_4 + \gamma \cdot X_4$, as in GLMBP.

\subsection{Multi-Receptive Design for Local Perception}
The DWConv modules in LMBP and GLMBP modules are configured with variable kernel sizes $K_{\text{sep}} \in \{3, 5, 7\}$ across different depth levels of the UltraLBM-UNet, achieving a dynamic multi-receptive field capability. Specifically, the kernel size $K_{\text{sep}}$ is assigned as follows, ensuring that deeper, lower-resolution layers capture wider local context:
\begin{itemize}
    \item Encoder: Stages IV, V, and VI use $K=3$, $K=5$, and $K=7$, respectively.
    \item Decoder: Stages I, II, and III use $K=7$, $K=5$, and $K=3$, respectively.
\end{itemize}
This design allows the network to effectively capture features at multiple spatial scales, improving the modeling of both fine-grained details and broader contextual information.
\subsection{Scaling Skip Connections}
We introduce a learnable scaling coefficient in each skip-connection. The rationale is twofold: first, different feature levels often contribute unequally to the final prediction, and uncontrolled skip-connection magnitudes can dominate or destabilize the decoding process. Second, medical images frequently exhibit large variations in texture and contrast, making it beneficial to adaptively adjust the strength of the information passed from the encoder.
Given a decoder feature map $\hat{X}_i$ and its corresponding encoder feature map $t_i$, the scaled skip-connection is defined as:
\begin{equation}
X_i = \hat{X}_i + k \cdot t_i,
\end{equation}
where $k \in \mathbb{R}$ is a learnable scalar shared across skip-connections.
By applying a scaling factor, the network can regulate the contribution of each skip-feature, leading to more stable optimization and more discriminative feature fusion.

\subsection{Distillation}

To further enhance the performance of extremely compact models without increasing inference complexity, we introduce a knowledge distillation (KD) strategy tailored for UltraLBM-UNet. Starting from the proposed architecture, a lightweight student model is constructed by uniformly reducing the channel dimensions of all stages by a factor of two, resulting in a substantially smaller parameter footprint while preserving the overall network topology.

The student model is trained under the supervision of the original full-capacity UltraLBM-UNet, which serves as the teacher. To effectively transfer structural, boundary, and spatial-focus knowledge critical for skin lesion segmentation, we design a hybrid distillation objective that combines standard supervision with teacher-guided regularization. The overall distillation loss is formulated as:
\begin{equation}
\mathcal{L}_{\text{distill}} = \lambda_h \mathcal{L}_{\text{hard}} + \lambda_s \mathcal{L}_{\text{DKD}} + \lambda_a \mathcal{L}_{\text{AT}} + \lambda_g \mathcal{L}_{\text{grad}},
\end{equation}
where $\mathcal{L}_{\text{hard}}$ denotes direct supervision from ground-truth annotations, while $\mathcal{L}_{\text{DKD}}$, $\mathcal{L}_{\text{AT}}$, and $\mathcal{L}_{\text{grad}}$ guide the student to align with the teacher's predictions across different feature dimensions.

\subsubsection{Semantic and Structural Alignment}
The student is optimized using the original segmentation loss with respect to the ground truth:
\begin{equation}
\mathcal{L}_{\text{hard}} = \mathcal{L}_{\text{seg}}(S, Y),
\end{equation}
where $S \in [0,1]^{H \times W}$ denotes the student prediction and $Y$ is the corresponding ground-truth mask. To transfer global structural knowledge, we adopt a Decoupled Knowledge Distillation (DKD) strategy \blue{~\cite{zhao2022decoupled}} that separately models foreground and background distributions:
{\small
\begin{equation}
\mathcal{L}_{\text{DKD}} = \mathbb{E}_{p \in \Omega} \Big[
T(p)\log\frac{T(p)}{S(p)} 
+ (1-T(p))\log\frac{1-T(p)}{1-S(p)}
\Big],
\end{equation}
}

where $T(p)$ and $S(p)$ denote the teacher and student prediction probabilities at pixel location $p$, respectively. This formulation enables class-balanced knowledge transfer and improves global structural consistency in the compressed model.

\subsubsection{Spatial Focus Transfer}
To preserve spatial focus and emphasize lesion regions, we introduce an Attention Transfer (AT) mechanism that aligns the spatial response distributions of the student and teacher. Given the flattened prediction maps, the spatial attention distributions are defined as:
\begin{equation}
\begin{split}
A_S &= \text{Softmax}(\text{vec}(S)/\tau_a), \\
A_T &= \text{Softmax}(\text{vec}(T)/\tau_a),
\end{split}
\end{equation}
where $\tau_a$ is a temperature parameter controlling the sharpness of the attention distribution. The attention transfer loss is then computed using Kullback--Leibler (KL) divergence:
\begin{equation}
\mathcal{L}_{\text{AT}} = \text{KL}(A_T \,\|\, A_S).
\end{equation}

\subsubsection{Boundary Fidelity Refinement}
To capture high-frequency edge information, which is often compromised during model compression, we implement a Gradient Matching Loss $\mathcal{L}_{\text{grad}}$. By applying Sobel operators $\nabla$, we align the gradient magnitude of the student's output with that of the teacher:
{
\small
\begin{equation}
\mathcal{L}_{\text{grad}} = \left\| \sqrt{(\nabla_x S)^2 + (\nabla_y S)^2} - \sqrt{(\nabla_x T)^2 + (\nabla_y T)^2} \right\|^2_2.
\end{equation}
}
By maintaining the same inference architecture and introducing distillation only during training, the proposed strategy enhances the representational capacity of the ultra-lightweight student model without additional runtime cost, making it well-suited for point-of-care deployment scenarios.

\section{Experiments}

\begin{table*}[t]
\centering
\caption{Comparison of DSC and IoU on ISIC2017, ISIC2018, and PH\textsuperscript{2} datasets, model complexity, and inference speed. $\uparrow \downarrow$ denotes the higher (lower) the better. $-$ means missing data from the source. \textbf{Bold} and \underline{underline} represent the best and the second best results. \blue{On-device latency and FPS are measured on a Xiaomi 2311DRK48C smartphone (Android~14, 4 CPU threads, ONNX Runtime, batch size~1, $256\times256$ input), averaged over 5 runs after 3 warmup runs. GPU latency and FPS are measured via PyTorch on an NVIDIA GPU (batch size~1, $256\times256$ input), averaged over 5 runs after 3 warmup runs.}}
\label{tab:combined_perf_complexity}
\resizebox{\textwidth}{!}{
\begin{tabular}{l c | cc | cc | cc | c c | c c | c c}
\toprule
\multirow{2}{*}{\textbf{Model}} & \multirow{2}{*}{\textbf{Venue (year)}}
& \multicolumn{2}{c|}{\textbf{ISIC2017}}
& \multicolumn{2}{c|}{\textbf{ISIC2018}}
& \multicolumn{2}{c|}{\textbf{PH\textsuperscript{2}}}
& \multirow{2}{*}{\textbf{Params (M)}$\downarrow$}
& \multirow{2}{*}{\textbf{FLOPs (G)}$\downarrow$}
& \multirow{2}{*}{\blue{\textbf{Latency (ms)}$\downarrow$}}
& \multirow{2}{*}{\blue{\textbf{FPS}$\uparrow$}}
& \multirow{2}{*}{\blue{\textbf{GPU Lat. (ms)}$\downarrow$}}
& \multirow{2}{*}{\blue{\textbf{GPU FPS}$\uparrow$}} \\
\cmidrule(lr){3-8}
& & \textbf{IoU}(\%)$\uparrow$ & \textbf{DSC}(\%)$\uparrow$
& \textbf{IoU}(\%)$\uparrow$ & \textbf{DSC}(\%)$\uparrow$
& \textbf{IoU}(\%)$\uparrow$ & \textbf{DSC}(\%)$\uparrow$
& & & & & & \\
\midrule

U-Net~\cite{ronneberger2015u} & MICCAI (2015) & 75.84 $\pm$ 1.26 & 86.25 $\pm$ 0.81 & 78.40 $\pm$ 0.53 & 87.89 $\pm$ 0.33 & 81.91 $\pm$ 1.14 & 90.05 $\pm$ 0.69 & 7.766 & 13.746 & \blue{212.400} & \blue{4.708} & \blue{11.279} & \blue{88.657} \\

UNet++~\cite{zhou2018unet++} & DLMIA (2018) & 76.64 $\pm$ 0.76 & 86.77 $\pm$ 0.49 & 78.38 $\pm$ 0.35 & 87.88 $\pm$ 0.22 & 82.65 $\pm$ 0.98 & 90.50 $\pm$ 0.59 & 9.163 & 34.903 & \blue{613.000} & \blue{1.631} & \blue{25.233} & \blue{39.631} \\

Attention U-Net~\cite{oktay2018attention} & MIDL (2018) & 77.02 $\pm$ 0.32 & 87.02 $\pm$ 0.21 & 78.46 $\pm$ 0.63 & 87.93 $\pm$ 0.40 & 83.22 $\pm$ 0.56 & 90.84 $\pm$ 0.33 & 34.879 & 66.632 & \blue{295.200} & \blue{3.388} & \blue{14.672} & \blue{68.158} \\

UNeXt~\cite{valanarasu2022unext} & MICCAI (2022) & 77.81 $\pm$ 0.65 & 87.52 $\pm$ 0.41 & 79.06 $\pm$ 0.19 & 88.30 $\pm$ 0.12 & 83.52 $\pm$ 1.09 & 91.02 $\pm$ 0.64 & 1.472 & 0.573 & \blue{38.000} & \blue{26.316} & \blue{\underline{5.116}} & \blue{\underline{195.447}} \\

MALUNet~\cite{ruan2022malunet} & BIBM (2022) & \underline{78.71 $\pm$ 0.84} & \underline{88.09 $\pm$ 0.53} & 79.42 $\pm$ 0.26 & 88.53 $\pm$ 0.16 & 83.83 $\pm$ 1.10 & 91.20 $\pm$ 0.65 & 0.178 & 0.083 & \blue{22.800} & \blue{43.860} & \blue{11.465} & \blue{87.221} \\

U-Lite~\cite{dinh20231m} & APSIPA (2023) & 77.95 $\pm$ 0.63 & 87.61 $\pm$ 0.40 & 78.75 $\pm$ 0.39 & 88.11 $\pm$ 0.24 & 82.82 $\pm$ 1.76 & 90.59 $\pm$ 1.05 & 0.878 & 0.757 & \blue{$-$} & \blue{$-$} & \blue{\textbf{4.776}} & \blue{\textbf{209.391}} \\

EGE-UNet~\cite{ruan2023ege} & MICCAI (2023) & 78.32 $\pm$ 0.74 & 87.84 $\pm$ 0.46 & 79.45 $\pm$ 0.33 & 88.55 $\pm$ 0.21 & 83.36 $\pm$ 0.87 & 90.93 $\pm$ 0.52 & 0.053 & 0.072 & \blue{28.400} & \blue{35.211} & \blue{17.972} & \blue{55.641} \\

CMUNeXt-S~\cite{tang2024cmunext} & ISBI (2024) & 77.87 $\pm$ 0.29 & 87.56 $\pm$ 0.18 & 78.66 $\pm$ 0.17 & 88.06 $\pm$ 0.11 & 82.89 $\pm$ 0.93 & 90.64 $\pm$ 0.56 & 0.418 & 1.090 & \blue{59.200} & \blue{16.892} & \blue{7.729} & \blue{129.387} \\

Rolling-UNet-S~\cite{liu2024rolling} & AAAI (2024) & 77.59 $\pm$ 0.42 & 87.38 $\pm$ 0.26 & 79.43 $\pm$ 0.17 & 88.53 $\pm$ 0.10 & 81.99 $\pm$ 2.30 & 90.09 $\pm$ 1.39 & 1.783 & 2.102 & \blue{79.200} & \blue{12.626} & \blue{29.329} & \blue{34.096} \\

TinyU-Net~\cite{chen2024tinyu} & MICCAI (2024) & 77.89 $\pm$ 0.56 & 87.57 $\pm$ 0.36 & 78.49 $\pm$ 0.63 & 87.95 $\pm$ 0.39 & 82.48 $\pm$ 0.87 & 90.40 $\pm$ 0.53 & 0.481 & 1.659 & \blue{1439.800} & \blue{0.695} & \blue{14.283} & \blue{70.014} \\

UltraLight VM-UNet~\cite{wu2024ultralight} & Patterns (2024) & 77.93 $\pm$ 0.38 & 87.59 $\pm$ 0.24 & 78.93 $\pm$ 0.20 & 88.23 $\pm$ 0.13 & 83.31 $\pm$ 1.35 & 90.89 $\pm$ 0.80 & 0.045 & 0.069 & \blue{21.000} & \blue{47.619} & \blue{21.969} & \blue{45.520} \\

MambaU-Lite~\cite{nguyen2025mambau} & ICISN (2025) & 77.32 $\pm$ 0.40 & 87.21 $\pm$ 0.26 & 78.68 $\pm$ 0.78 & 88.07 $\pm$ 0.49 & 82.93 $\pm$ 0.98 & 90.66 $\pm$ 0.58 & 0.416 & 0.934 & \blue{$-$} & \blue{$-$} & \blue{41.419} & \blue{24.143} \\

U-KAN~\cite{li2024ukanmakesstrongbackbone} & AAAI (2025) & 77.71 $\pm$ 0.46 & 87.46 $\pm$ 0.29 & \underline{79.59 $\pm$ 0.32} & \underline{88.64 $\pm$ 0.20} & 84.02 $\pm$ 0.41 & 91.31 $\pm$ 0.25 & 25.359 & 6.889 & \blue{367.600} & \blue{2.720} & \blue{26.968} & \blue{37.081} \\

MK-UNet~\cite{rahman2025mk} & ICCVW (2025) & 78.57 $\pm$ 0.45 & 88.00 $\pm$ 0.28 & 79.45 $\pm$ 0.40 & 88.55 $\pm$ 0.25 & 83.85 $\pm$ 0.75 & 91.22 $\pm$ 0.44 & 0.316 & 0.328 & \blue{42.200} & \blue{23.697} & \blue{17.181} & \blue{58.202} \\

\midrule
\rowcolor{gray!10}
\textbf{UltraLBM-UNet-T (Ours)} & - & 78.57 $\pm$ 0.42 & 88.00 $\pm$ 0.27 & 78.82 $\pm$ 0.89 & 88.15 $\pm$ 0.56 & \textbf{84.92 $\pm$ 0.73} & \textbf{91.85 $\pm$ 0.42} & \textbf{0.011} & \textbf{0.019} & \blue{\textbf{18.200}} & \blue{\textbf{54.945}} & \blue{17.765} & \blue{56.290} \\
\rowcolor{gray!10}
\textbf{UltraLBM-UNet (Ours)} & - & \textbf{79.82 $\pm$ 0.73} & \textbf{88.78 $\pm$ 0.45} & \textbf{79.94 $\pm$ 0.64} & \textbf{88.85 $\pm$ 0.39} & \underline{84.41 $\pm$ 0.87} & \underline{91.54 $\pm$ 0.51} & \underline{0.034} & \underline{0.060} & \blue{\underline{20.800}} & \blue{\underline{48.077}} & \blue{16.563} & \blue{60.377} \\

\bottomrule
\end{tabular}}
\begin{tablenotes}
\item[*] Results on PH\textsuperscript{2} are averaged from six models trained separately on ISIC2017 and ISIC2018.
\item[*] \blue{U-Lite uses dilated convolutions with SAME padding, which is not supported by ONNX Runtime on Android.}
\item[*] \blue{MambaU-Lite contains dynamic pooling operations that cannot be resolved to static shapes during ONNX export.}
\end{tablenotes}
\end{table*}

\subsection{Datasets}
To evaluate the performance of our proposed model for skin lesion segmentation, two established public datasets are chosen for training: ISIC2017~\cite{codella2018skin} and ISIC2018~\cite{codella2019skin,tschandl2018ham10000}. The ISIC 2017 collection comprises a total of 2,150 dermoscopic images with corresponding masks, while the ISIC 2018 collection contains 2,694 images with corresponding masks. Following standard practice in previous studies, both datasets are randomly divided into a training set and a testing set, utilizing a 7:3 split ratio~\cite{ruan2022malunet}. In addition, the PH\textsuperscript{2} dataset~\cite{mendoncca2015ph2} is employed as an external validation set to assess the generalization ability of our model. All 200 images in PH\textsuperscript{2} are used for evaluation. Specifically, six models trained separately on ISIC 2017 and ISIC 2018 are applied to the PH\textsuperscript{2} images, and the reported results represent the average performance of these models.

\blue{To evaluate robustness against real-world image quality degradation, we additionally construct a degraded version of the PH\textsuperscript{2} test set simulating handheld smartphone-based acquisition under adverse conditions. Each image is sequentially perturbed by global brightness attenuation (factor 0.4), additive Gaussian noise ($\sigma=40$), and horizontal motion blur ($11\times11$ kernel), corresponding to low-light, sensor noise, and hand tremor, respectively. Segmentation masks are kept intact.}

\blue{To assess generalization beyond dermoscopy, we further evaluate UltraLBM-UNet on the BUSI breast ultrasound dataset~\cite{aldhabyani2020busi}, which differs in both anatomical region and imaging modality. BUSI contains 780 ultrasound images of benign, malignant, and normal cases with corresponding lesion masks, and is randomly split into training and test sets at an 8:2 ratio.}

\subsection{Implementation Details}
\label{sec:implementation}
All experiments are implemented in~\cite{paszke2019pytorch} and conducted on an NVIDIA Tesla T4 GPU (16\,GB). Images are resized to $256 \times 256$ pixels, normalized, and augmented via random flips and rotations, with identical transformations applied to segmentation masks. The model is trained for 300 epochs using the AdamW optimizer (learning rate $0.001$, weight decay $0.01$)~\cite{loshchilov2017decoupled} and a CosineAnnealingLR scheduler ($t_{\text{max}}=50$, $\eta_{\text{min}}=1\times10^{-5}$)~\cite{loshchilov2016sgdr} with a batch size of 8. Performance is assessed using IoU and DSC, averaged over three runs with different random seeds. In our experiments, we adapt a combined loss that integrates both BCE and Dice losses:
\blue{\begin{equation}\mathcal{L}=  BCE(\hat{y}, y)+ Dice(\hat{y}, y).\end{equation}}
For distillation experiments, the student model is constructed by uniformly halving the channel width of all stages in UltraLBM-UNet, while the pre-trained original model is used as the teacher. \blue{For the BUSI generalization experiment, all training hyperparameters follow the same settings as the ISIC experiments.} \blue{For the PH\textsuperscript{2} robustness evaluation, the checkpoints used in Table~\ref{tab:combined_perf_complexity} are applied directly to the degraded inputs without any retraining or fine-tuning.}

\blue{For inference speed evaluation, all models are exported to ONNX format (opset~12) and benchmarked under two settings, both at $256\times256$ resolution with batch size~1, averaged over 5 runs following 3 warmup runs. For on-device evaluation, models are deployed on a consumer smartphone (Xiaomi 2311DRK48C, Android~14) via ONNX Runtime with 4~CPU threads on 100 images randomly sampled from the ISIC 2017 test set. GPU latency is measured via PyTorch on the NVIDIA Tesla T4 GPU used for training. MambaU-Lite and U-Lite are excluded from on-device evaluation due to ONNX export incompatibilities: U-Lite uses dilated convolutions with SAME padding unsupported by ONNX Runtime on Android, while MambaU-Lite contains dynamic pooling operations that cannot be resolved to static shapes during ONNX export.}

\subsection{Results}
\begin{figure*}[h]
  \centering
  \includegraphics[width=0.98\linewidth]{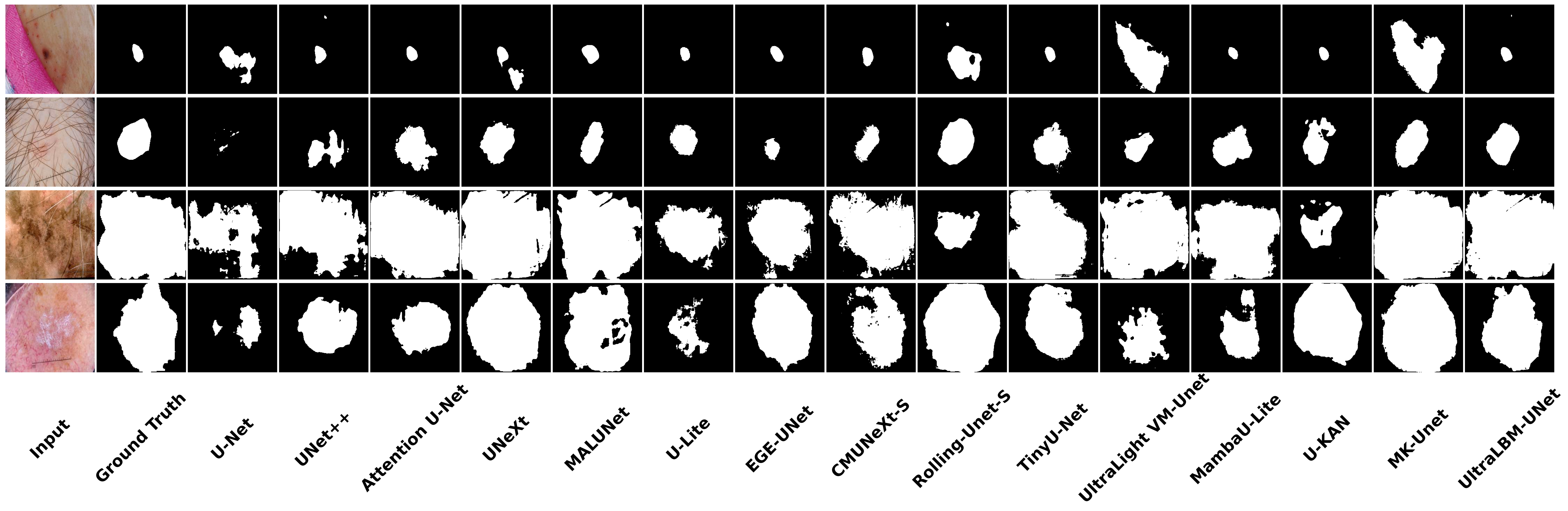}
  \caption{Comparison of segmentation results across representative ISIC 2017 cases.}
  \label{fig:Comparison_segmentation_results}
\end{figure*}
Fig.~\ref{fig:Comparison_segmentation_results} illustrates the robustness of UltraLBM-UNet and its training variant (UltraLBM-UNet-T) across diverse lesion appearances, where the models maintain consistent mask shapes and semantic completeness under varying color, texture, and boundary complexity.

Quantitative results are presented in Table~\ref{tab:combined_perf_complexity}, which compares UltraLBM-UNet and UltraLBM-UNet-T against recent state-of-the-art methods in terms of segmentation accuracy (IoU and DSC) and model complexity (Params and FLOPs) on the ISIC2017, ISIC2018, and PH\textsuperscript{2} datasets. UltraLBM-UNet achieves the best overall accuracy–efficiency trade-off among lightweight models, delivering top performance on ISIC2017 and ISIC2018 with minimal computational cost. Under the same inference complexity constraints, the distilled variant UltraLBM-UNet-T further improves segmentation accuracy in the ultra-lightweight regime, achieving superior performance on PH\textsuperscript{2} despite a substantially reduced parameter budget. \blue{Beyond static complexity metrics, the reported on-device inference experiments further confirm that the efficiency advantages of UltraLBM-UNet translate directly into practical deployment speed, with both proposed variants achieving the lowest inference latency among all compared methods on a consumer smartphone.}

\subsubsection{Comparison with SOTA Methods}

UltraLBM-UNet and UltraLBM-UNet-T exhibit high accuracy and strong competitiveness across both benchmarks:
\begin{itemize}
    \item \textbf{ISIC 2017:} UltraLBM-UNet achieves the highest mean IoU ($79.82\%$) and DSC ($88.78\%$), establishing SOTA performance by surpassing all compared methods.
    \item \textbf{ISIC 2018:} UltraLBM-UNet achieves strong performance with an IoU of ~$79.94\%$ and a DSC of ~$88.85\%$, achieving the best performance compared to SOTA methods.
    \item \textbf{PH\textsuperscript{2}:} UltraLBM-UNet-T attains the highest IoU ($84.92 \%$) and DSC ($91.85 \%$), demonstrating excellent generalization on this small, high-quality dermoscopic dataset. UltraLBM-UNet achieves the second best.
\end{itemize}

\subsubsection{Model Efficiency Analysis}
A key advantage of the proposed framework lies in its scalability across different efficiency regimes. UltraLBM-UNet achieves an excellent balance between accuracy and efficiency, requiring only $0.034$\,M parameters and $0.060$\,G FLOPs. For more constrained deployment scenarios, the distilled variant UltraLBM-UNet-T further reduces the model size to $0.011$\,M parameters and $0.019$\,G FLOPs, while retaining competitive segmentation accuracy.

Compared with existing lightweight architectures such as MALUNet, EGE-UNet, and UltraLight VM-UNet, both UltraLBM-UNet and UltraLBM-UNet-T exhibit substantially lower computational costs, highlighting their suitability for point-of-care and resource-limited clinical applications.

\subsubsection{On-Device and GPU Inference Speed}
\blue{Table~\ref{tab:combined_perf_complexity} reports both on-device and GPU inference speed. On mobile, UltraLBM-UNet-T achieves the lowest latency of 18.2\,ms (54.9\,FPS), followed by UltraLBM-UNet at 20.8\,ms (48.1\,FPS), both outperforming all compared methods. The closest competitors, UltraLight VM-UNet (21.0\,ms) and MALUNet (22.8\,ms), remain slower despite comparable or larger parameter budgets. On GPU, U-Lite achieves the fastest inference (4.8\,ms, 209.4\,FPS), followed by UNeXt (5.1\,ms, 195.4\,FPS). Our models rank competitively at 16.6\,ms and 17.8\,ms on GPU, on par with MK-UNet and EGE-UNet while achieving substantially better segmentation accuracy. Notably, TinyU-Net exhibits the worst on-device latency (1439.8\,ms) among all tested models, and CMUNeXt-S, despite being fast on GPU (7.7\,ms), degrades to 59.2\,ms on mobile, demonstrating that GPU speed does not reliably predict deployment efficiency on resource-constrained devices.}

\blue{The superior on-device speed of our models is consistent with their extremely low parameter count (0.034M for UltraLBM-UNet and 0.011M for UltraLBM-UNet-T) and FLOPs (0.060G and 0.019G respectively), which are the lowest among all compared methods. In addition, the identity shortcut in the GLMBP module introduces zero additional computation, further reducing per-inference overhead. These results confirm that the efficiency of UltraLBM-UNet translates directly into practical mobile deployment speed, and that FLOPs and parameter counts alone are insufficient proxies for on-device performance.}

\subsubsection{\blue{Robustness to Real-World Acquisition Degradation}}

\blue{Table~\ref{tab:robustness_ph2} reports segmentation performance on the degraded 
PH\textsuperscript{2} test set, with U-Net included as a representative reference baseline. Under the compounded low-light, noise, and motion-blur degradation, UltraLBM-UNet retains 80.92\% IoU and 89.44\% DSC, substantially outperforming 
U-Net (75.04\% / 85.66\%) despite using more than 200$\times$ fewer parameters, with absolute IoU and DSC drops of 3.49\% and 2.10\%, respectively, compared to 6.87\% and 4.39\% for U-Net. The standard deviation of IoU across the six checkpoints decreases from 5.16 (U-Net) to 1.74 (UltraLBM-UNet), representing a 2.97-fold reduction in prediction variance, which suggests that bidirectional global context modeling contributes to more stable predictions under adverse acquisition conditions that are not readily achievable through purely local convolutional aggregation. These results provide direct empirical evidence that the efficiency of UltraLBM-UNet does not come at the cost of robustness, supporting its suitability for real-world point-of-care deployment.}

\blue{\begin{table}[t]
\centering
\scriptsize
\setlength{\tabcolsep}{4pt}
\caption{Robustness evaluation on PH\textsuperscript{2} under severe acquisition 
degradation. $\Delta$ denotes the absolute drop relative to clean PH\textsuperscript{2}. }
\label{tab:robustness_ph2}
\resizebox{\columnwidth}{!}{
\begin{tabular}{l|cc|cc|cc}
\toprule
\multirow{2}{*}{\textbf{Model}} 
& \multicolumn{2}{c|}{\textbf{PH\textsuperscript{2} (clean)}} 
& \multicolumn{2}{c|}{\textbf{PH\textsuperscript{2} (degraded)}} 
& \multicolumn{2}{c}{\textbf{$\Delta$ (\%)}} \\
\cmidrule(lr){2-3} \cmidrule(lr){4-5} \cmidrule(lr){6-7}
& \textbf{IoU}(\%)$\uparrow$ & \textbf{DSC}(\%)$\uparrow$
& \textbf{IoU}(\%)$\uparrow$ & \textbf{DSC}(\%)$\uparrow$
& \textbf{IoU}$\downarrow$ & \textbf{DSC}$\downarrow$ \\
\midrule
U-Net~\cite{ronneberger2015u} 
& 81.91 $\pm$ 1.14 & 90.05 $\pm$ 0.69 
& 75.04 $\pm$ 5.16 & 85.66 $\pm$ 3.47 
& $-$6.87 & $-$4.39 \\
\rowcolor{gray!10}
\textbf{UltraLBM-UNet (Ours)} 
& \textbf{84.41 $\pm$ 0.87} & \textbf{91.54 $\pm$ 0.51} 
& \textbf{80.92 $\pm$ 1.74} & \textbf{89.44 $\pm$ 1.07} 
& \textbf{$-$3.49} & \textbf{$-$2.10} \\
\bottomrule
\end{tabular}}
\end{table}}

\begin{table}[t]
\centering
\scriptsize
\setlength{\tabcolsep}{6pt}
\caption{\blue{Generalization evaluation on the BUSI breast ultrasound dataset. 
\textbf{Bold} and \underline{underline} indicate the best and second-best results.}}
\label{tab:busi_generalization}
\begin{tabular}{l|cc}
\toprule
\textbf{Model} & \textbf{IoU}(\%)$\uparrow$ & \textbf{DSC}(\%)$\uparrow$ \\
\midrule
UltraLight VM-UNet~\cite{wu2024ultralight}    & 53.28 & 69.52 \\
EGE-UNet~\cite{ruan2023ege}                   & 57.25 & 72.81 \\
MALUNet~\cite{ruan2022malunet}                & \underline{58.21} & \underline{73.59} \\
\rowcolor{gray!10}
\textbf{UltraLBM-UNet (Ours)}                 & \textbf{59.39} & \textbf{74.53} \\
\bottomrule
\end{tabular}
\end{table}
\subsubsection{\blue{Generalization to Breast Ultrasound Segmentation}}

\blue{To examine whether the proposed global–local design generalizes beyond dermoscopy, we further evaluate UltraLBM-UNet on the BUSI breast ultrasound dataset, comparing against representative lightweight baselines at a comparable parameter and computational scale. As shown in Table~\ref{tab:busi_generalization}, UltraLBM-UNet achieves the highest IoU (59.39\%) and DSC (74.53\%), outperforming the second-best method (MALUNet) by 1.18\% IoU and 0.94\% DSC. The absolute accuracy on BUSI remains lower than on the dermoscopic benchmarks, reflecting the intrinsic difficulty of ultrasound segmentation under speckle noise, low contrast, and limited training data; nevertheless, UltraLBM-UNet consistently leads its lightweight counterparts under identical conditions, indicating that the proposed design generalizes across 2D medical imaging modalities while leaving room for further improvement in the ultrasound domain.}

\subsubsection{Ablation Study}
\begin{table}[h]
\centering
\scriptsize
\caption{Ablation study on long-range dependency modeling (ISIC2017 dataset).}
\label{tab:ablation_a}
\begin{tabular*}{\columnwidth}{@{\extracolsep{\fill}}l|c|c|c|c@{}}
\toprule
Model & Params $\downarrow$ & FLOPs $\downarrow$ & IoU $\uparrow$ & DSC $\uparrow$ \\
 & (M) & (G) & (\%) & (\%) \\
\midrule
All GLMBP & 0.036 & 0.063 & 79.27 & 88.43 \\
All LMBP & 0.036 & 0.063 & 78.86 & 88.18 \\
2 LMBP + GLMBP 
& \textbf{0.032} & \textbf{0.060} & 79.06 & 88.30 \\
\rowcolor{gray!10}
\textbf{LMBP + 2 GLMBP  (Ours)} 
& 0.034 & 0.060 & \textbf{79.82} & \textbf{88.78} \\
\bottomrule
\end{tabular*}
\end{table}

\begin{table}[h]
\centering
\scriptsize
\setlength{\tabcolsep}{3pt}
\renewcommand{\arraystretch}{1.05}
\caption{Ablation study on GLMBP multi-branch decomposition (ISIC2017 dataset).}
\label{tab:ablation_b}
\begin{tabular*}{\columnwidth}{@{\extracolsep{\fill}}l|c|c|c|c@{}}
\toprule
Model & Params $\downarrow$ & FLOPs $\downarrow$ & IoU $\uparrow$ & DSC $\uparrow$ \\
 & (M) & (G) & (\%) & (\%) \\
\midrule
Single-Direction Mamba & 0.034 & 0.060 & 78.65 & 88.05 \\

Bi-Mamba (Unshared Weights) &0.043& 0.060&78.19 & 87.76 \\
 
 Bi-Self-Attention & \textbf{0.028} & \textbf{0.059} & 78.25 & 87.80 \\
2 DWConv + 1 Bi-Mamba + Identity & 0.037 & 0.060 & 78.39 & 87.88 \\
2 DWConv + 2 Bi-Mamba & 0.043 & 0.060 & 78.31 & 87.83 \\
1 DWConv + 3 Bi-Mamba & 0.034 & 0.061 & 78.31 & 87.80 \\
2 Identity + 2 Bi-Mamba  &  0.031 & 0.061 & 78.25 & 87.84 \\
\rowcolor{gray!10}
\textbf{1 DWConv + 2 Bi-Mamba + Identity} 
&0.034 & 0.060 & \textbf{79.82} & \textbf{88.78} \\
\bottomrule
\end{tabular*}
\end{table}

\begin{table}[h]
\centering
\scriptsize
\setlength{\tabcolsep}{3pt}
\renewcommand{\arraystretch}{1.05}
\caption{Ablation study on LMBP multi-branch decomposition (ISIC2017 dataset).}
\label{tab:ablation_c}
\begin{tabular*}{\columnwidth}{@{\extracolsep{\fill}}l|c|c|c|c@{}}
\toprule
Model & Params $\downarrow$ & FLOPs $\downarrow$ & IoU $\uparrow$ & DSC $\uparrow$ \\
 & (M) & (G) & (\%) & (\%) \\
\midrule
4-DWConv (LMBP) & 0.037 & 0.062 & 78.81 & 88.15 \\
Single-Branch Conv ($k$=3) & 0.045 & 0.067 & 78.55 & 87.99 \\
\rowcolor{gray!10}
\textbf{3-DWConv + Identity (Ours) } 
& \textbf{0.034} & \textbf{0.060} & \textbf{79.82} & \textbf{88.78} \\
\bottomrule
\end{tabular*}
\end{table}

\begin{table}[h]
\centering
\scriptsize
\caption{Ablation study on kernel strategy (ISIC2017 dataset).}
\label{tab:ablation_f}
\begin{tabular*}{\columnwidth}{@{\extracolsep{\fill}}l|c|c|c|c@{}}
\toprule
Model & Params $\downarrow$ & FLOPs $\downarrow$ & IoU $\uparrow$ & DSC $\uparrow$ \\
 & (M) & (G) & (\%) & (\%) \\
\midrule
Unified Kernel = 7 & 0.036 & 0.062 & 79.18 & 88.38 \\
Unified Kernel = 5 & 0.033 & 0.061 & 78.80 & 88.14 \\
Unified Kernel = 3 & \textbf{0.033} & \textbf{0.060} & 78.66 & 88.06 \\
Kernel = 3, 3, 5 & 0.033 & 0.060 & 78.80 & 88.14 \\
Kernel = 3, 5, 5 & 0.033 & 0.060 & 78.35 & 87.86 \\
\rowcolor{gray!10}
\textbf{Kernel = 3, 5, 7 (Ours)} 
& 0.034 & 0.060 & \textbf{79.82} & \textbf{88.78} \\
\bottomrule
\end{tabular*}
\end{table}

\paragraph{(a) Combination of GLMBP and LMBP modules.}

Table~\ref{tab:ablation_a} investigates the effect of the placement of GLMBP and LMBP modules. Replacing GLMBP with LMBP at all stages, thereby removing global modeling entirely, leads to a clear performance drop (from 79.27\% to 78.86\% IoU) under identical parameter and FLOP budgets, indicating that purely local convolutional operations are insufficient for capturing lesion-level structural coherence.

When local and global modules are combined, configurations with increased GLMBP usage consistently achieve higher segmentation accuracy. In particular, the proposed design with multiple GLMBP stages outperforms the single-GLMBP variant by 0.76\% IoU under comparable computational cost, confirming that bidirectional global dependency modeling is essential for aggregating long-range context and modeling the smooth, spatially continuous lesion structures commonly observed in dermoscopic images.

\paragraph{(b) Multi-Branch Decomposition and Information Routing in GLMBP.}
Table~\ref{tab:ablation_b} evaluates how different configurations of the four-branch GLMBP module affect segmentation performance by varying the allocation of global, local, and identity branches. All ablated variants underperform the full UltraLBM-UNet, confirming that the original composition of two Bi-Mamba branches, one DWConv-based local branch, and one identity pathway provides the most effective feature routing strategy.

Replacing the bidirectional Mamba with a single-direction variant results in a clear performance degradation, highlighting the importance of symmetric global information propagation. Similarly, unsharing weights between the forward and backward Mamba branches increases model complexity while further reducing accuracy, indicating that weight sharing is critical for both efficiency and coherent global modeling. Although Bi-Self-Attention achieves the lowest parameter and FLOP cost, it consistently underperforms Mamba-based configurations, suggesting that state-space recurrence is better suited for capturing long-range dependencies in dermoscopic images under strict efficiency constraints.

Adjusting the balance between global and local processing also leads to degraded performance. Increasing the number of DWConv branches or over-emphasizing Bi-Mamba branches (e.g., replacing the identity path or local branch) consistently harms segmentation accuracy, demonstrating that unbalanced information routing disrupts effective feature fusion. In contrast, the proposed configuration maintains a well-calibrated interaction between global context aggregation, local detail preservation, and identity-based feature reuse, which none of the alternative decompositions can match.

\paragraph{(c) Local Branch Decomposition in LMBP.}
Table~\ref{tab:ablation_c} examines the impact of different local multi-branch configurations within the LMBP module.  Replacing the proposed LMBP design with either four DWConv branches or a single convolutional branch consistently degrades segmentation performance while increasing model complexity. In contrast, the proposed configuration that combines three DWConv branches with an identity pathway achieves the best accuracy with the lowest parameter and FLOP cost. This result indicates that purely local convolutional processing is insufficient for effective feature fusion, and that preserving an identity branch plays a critical role in maintaining fine-grained spatial information and stable feature propagation within the local perception module.

\paragraph{(d) Kernel Strategy for Multi-Receptive Local Perception.}
Table~\ref{tab:ablation_f} compares different kernel assignment strategies to evaluate the effect of depth-aware receptive field adaptation. While using a unified kernel size across all layers yields reasonable performance, it consistently underperforms the proposed progressive strategy. In particular, larger kernels improve coarse spatial aggregation but fail to preserve fine boundary details in shallow layers, whereas smaller kernels limit the ability of deeper layers to capture large-scale lesion morphology. In contrast, the progressive kernel assignment ($3\!\rightarrow\!5\!\rightarrow\!7$ in the encoder and $7\!\rightarrow\!5\!\rightarrow\!3$ in the decoder) achieves the best segmentation accuracy under comparable computational cost, demonstrating that dermoscopic lesion structures benefit from receptive fields that adapt with semantic depth rather than from a uniform spatial scale.

\paragraph{(e) Scaling in Skip-Connections.}
Table~\ref{tab:ablation_d} presents the effect of introducing a scaling factor in skip-connections on the ISIC2017 dataset. Adding the scaling leads to an improvement in segmentation performance, with IoU increasing from 79.24\% to 79.82\% and DSC from 88.42\% to 88.78\%, while the number of parameters and FLOPs remain almost unchanged. This indicates that the learnable scaling effectively modulates the contribution of features propagated through skip-connections, enhancing feature fusion between encoder and decoder without additional computational cost. In summary, scaling in skip-connections improves the network's ability to integrate multi-level features, resulting in more accurate lesion delineation.

\begin{table}[h]
\centering
\scriptsize
\caption{Ablation study on scaling in skip-connections (ISIC2017 dataset).}
\label{tab:ablation_d}
\begin{tabular*}{\columnwidth}{@{\extracolsep{\fill}}l|c|c|c|c@{}}
\toprule
Skip Scaling& Params $\downarrow$ & FLOPs $\downarrow$ & IoU $\uparrow$ & DSC $\uparrow$ \\
 & (M) & (G) & (\%) & (\%) \\
\midrule
w/o & \textbf{0.034}&\textbf{0.060}&79.24&88.42 \\
Stage-wise & 0.034&0.060&78.52&87.97 \\
w & 0.034 & 0.060 & \textbf{79.82} & \textbf{88.78} \\
\bottomrule
\end{tabular*}
\end{table}

\begin{table}[h]
\centering
\scriptsize
\setlength{\tabcolsep}{3pt}
\renewcommand{\arraystretch}{1.05}
\caption{Ablation study on Distillation Strategy (ISIC2017 dataset).}
\label{tab:ablation_g}
\begin{tabular*}{\columnwidth}{@{\extracolsep{\fill}}l|c|c|c|c@{}}
\toprule
Model & Params $\downarrow$ & FLOPs $\downarrow$ & IoU $\uparrow$ & DSC $\uparrow$ \\
 & (M) & (G) & (\%) & (\%) \\
\midrule
w/o Distill&0.011&0.019&77.30&87.20\\
KL-Distill&0.011&0.019&77.39&87.25\\
DKD-Distill&0.011&0.019&78.19&87.76\\
DKD+Att &0.011&0.019&78.25&87.80\\
DKD+Grad&0.011&0.019&77.91&87.58\\
\rowcolor{gray!10}
\textbf{Full Strategy } & 0.011 & 0.019 & \textbf{78.57} & \textbf{88.00}\\
\bottomrule
\end{tabular*}
\end{table}

\paragraph{(g) Effect of Distillation Strategy.}
Table~\ref{tab:ablation_g} presents an ablation study on the proposed distillation strategy using the ultra-lightweight student model. Training the student without distillation results in the lowest segmentation accuracy, indicating that severe channel reduction significantly limits representational capacity. Introducing conventional KL-based distillation yields only marginal improvements, suggesting that naive probabilistic alignment is insufficient for effective knowledge transfer in this setting.

In contrast, the decoupled knowledge distillation (DKD) strategy provides a notable performance gain, demonstrating its effectiveness in transferring class-balanced structural information from the teacher. Adding gradient-based matching alone does not further improve performance and slightly degrades accuracy, indicating that direct gradient alignment may introduce instability in the ultra-light regime. The full distillation strategy, which jointly integrates decoupled probabilistic distillation, gradient-based boundary supervision, and spatial attention transfer, achieves the best performance under identical computational cost. These results confirm that effective distillation for ultra-compact segmentation models requires both balanced probabilistic guidance and explicit spatial focus alignment, rather than a single distillation component.

\section{Conclusion}

In this work, we introduce UltraLBM-UNet, a lightweight yet highly effective skin lesion segmentation model that integrates a bidirectional Mamba-based global modeling mechanism with multi-branch local feature perception. Evaluations on ISIC 2017, ISIC 2018 and PH\textsuperscript{2} demonstrate that UltraLBM-UNet consistently delivers state-of-the-art segmentation accuracy, outperforming existing lightweight and Mamba-based counterparts with the lowest parameter count and FLOPs. \blue{Ablation studies further reveal a consistent architectural principle underlying the proposed design: adaptive modulation mechanisms, such as weight-shared bidirectional Mamba and learnable skip-connection scaling, consistently outperform independently parameterized alternatives, demonstrating that deliberate structural choices, rather than increased parameter budgets, are the key to effective ultra-lightweight segmentation.} \blue{On-device inference experiments additionally confirm that UltraLBM-UNet achieves the fastest mobile inference speed among all compared methods, validating its practical suitability for real-time point-of-care deployment.} These results establish UltraLBM-UNet as a principled and deployable solution for resource-constrained clinical environments, where accurate, robust, and efficient lesion analysis is essential.

\paragraph{Limitations and Future Works}

Although UltraLBM-UNet demonstrates strong segmentation performance with extremely low parameter and FLOP budgets, several limitations remain. \blue{First, while the present evaluation has been extended from dermoscopic images to breast ultrasound (Table~\ref{tab:busi_generalization}) to demonstrate generalization across 2D medical imaging modalities, volumetric modalities such as CT and MRI remain to be investigated; adapting the sequence flattening strategy and adopting hybrid 2D--3D state-space formulations would likely be necessary to preserve volumetric spatial priors.} Second, while the bidirectional Mamba mechanism effectively captures long-range dependencies, its sequence-based formulation still requires flattening 2D feature maps, which may lead to limited preservation of spatial priors when handling highly irregular lesion shapes. \blue{Third, while we present a preliminary robustness evaluation on a degraded version of PH\textsuperscript{2} simulating low-light, sensor noise, and motion blur (Table~\ref{tab:robustness_ph2}), this synthetic degradation pipeline does not fully capture the diversity of real-world acquisition artifacts encountered in clinical practice, such as specular reflections, focus drift, or scale variation. A systematic robustness study on real handheld or smartphone-acquired dermoscopic images remains an important direction for future clinical validation.}

Future work will explore the extension of UltraLBM-UNet to multimodal dermatology datasets, cross-domain skin lesion collections \blue{and additional medical imaging modalities including ultrasound, CT, and MRI}. \blue{For volumetric modalities, improving the spatial fidelity of Mamba-based modules through hybrid 2D--3D state-space formulations or grid-aware recurrence mechanisms will also be investigated, with the goal of enabling robust, resource-efficient segmentation across diverse clinical imaging environments.}

\printcredits

\section*{Declaration of Generative AI and AI-assisted Technologies in the Manuscript Preparation Process}

During the preparation of this manuscript, we used ChatGPT (OpenAI) to assist with language refinement and clarity improvement. After using this tool, we carefully reviewed, edited, and verified all content to ensure accuracy and originality, and we take full responsibility for the content of the published article.

\bibliographystyle{elsarticle-num}
\bibliography{reference}

@inproceedings{cao2022swin,
  title={Swin-unet: Unet-like pure transformer for medical image segmentation},
  author={Cao, Hu and Wang, Yueyue and Chen, Joy and Jiang, Dongsheng and Zhang, Xiaopeng and Tian, Qi and Wang, Manning},
  booktitle={European conference on computer vision},
  pages={205--218},
  year={2022},
  organization={Springer}
}

@article{chen2021transunet,
  title={Transunet: Transformers make strong encoders for medical image segmentation},
  author={Chen, Jieneng and Lu, Yongyi and Yu, Qihang and Luo, Xiangde and Adeli, Ehsan and Wang, Yan and Lu, Le and Yuille, Alan L and Zhou, Yuyin},
  journal={arXiv preprint arXiv:2102.04306},
  year={2021}
}

@article{dosovitskiy2020image,
  title={An image is worth 16x16 words: Transformers for image recognition at scale},
  author={Dosovitskiy, Alexey and Beyer, Lucas and Kolesnikov, Alexander and Weissenborn, Dirk and Zhai, Xiaohua and Unterthiner, Thomas and Dehghani, Mostafa and Minderer, Matthias and Heigold, Georg and Gelly, Sylvain and others},
  journal={arXiv preprint arXiv:2010.11929},
  year={2020}
}

@article{hu2024leanet,
  title={LeaNet: Lightweight U-shaped architecture for high-performance skin cancer image segmentation},
  author={Hu, Binbin and Zhou, Pan and Yu, Hongfang and Dai, Yueyue and Wang, Ming and Tan, Shengbo and Sun, Ying},
  journal={Computers in Biology and Medicine},
  volume={169},
  pages={107919},
  year={2024},
  publisher={Elsevier}
}

@inproceedings{huang2020unet,
  title={Unet 3+: A full-scale connected unet for medical image segmentation},
  author={Huang, Huimin and Lin, Lanfen and Tong, Ruofeng and Hu, Hongjie and Zhang, Qiaowei and Iwamoto, Yutaro and Han, Xianhua and Chen, Yen-Wei and Wu, Jian},
  booktitle={ICASSP 2020-2020 IEEE international conference on acoustics, speech and signal processing (ICASSP)},
  pages={1055--1059},
  year={2020},
  organization={Ieee}
}

@article{li2024dseunet,
  title={DSEUNet: A lightweight UNet for dynamic space grouping enhancement for skin lesion segmentation},
  author={Li, Jian and Wang, Jiawei and Lin, Fengwu and Wu, Wenqi and Chen, Zhao-Min and Heidari, Ali Asghar and Chen, Huiling},
  journal={Expert Systems with Applications},
  volume={255},
  pages={124544},
  year={2024},
  publisher={Elsevier}
}

@inproceedings{milletari2016v,
  title={V-net: Fully convolutional neural networks for volumetric medical image segmentation},
  author={Milletari, Fausto and Navab, Nassir and Ahmadi, Seyed-Ahmad},
  booktitle={2016 fourth international conference on 3D vision (3DV)},
  pages={565--571},
  year={2016},
  organization={Ieee}
}

@article{oktay2018attention,
  title={Attention u-net: Learning where to look for the pancreas},
  author={Oktay, Ozan and Schlemper, Jo and Folgoc, Loic Le and Lee, Matthew and Heinrich, Mattias and Misawa, Kazunari and Mori, Kensaku and McDonagh, Steven and Hammerla, Nils Y and Kainz, Bernhard and others},
  journal={arXiv preprint arXiv:1804.03999},
  year={2018}
}

@inproceedings{ronneberger2015u,
  title={U-net: Convolutional networks for biomedical image segmentation},
  author={Ronneberger, Olaf and Fischer, Philipp and Brox, Thomas},
  booktitle={International Conference on Medical image computing and computer-assisted intervention},
  pages={234--241},
  year={2015},
  organization={Springer}
}

@inproceedings{ruan2022malunet,
  title={Malunet: A multi-attention and light-weight unet for skin lesion segmentation},
  author={Ruan, Jiacheng and Xiang, Suncheng and Xie, Mingye and Liu, Ting and Fu, Yuzhuo},
  booktitle={2022 IEEE International Conference on Bioinformatics and Biomedicine (BIBM)},
  pages={1150--1156},
  year={2022},
  organization={IEEE}
}

@inproceedings{ruan2023ege,
  title={Ege-unet: an efficient group enhanced unet for skin lesion segmentation},
  author={Ruan, Jiacheng and Xie, Mingye and Gao, Jingsheng and Liu, Ting and Fu, Yuzhuo},
  booktitle={International conference on medical image computing and computer-assisted intervention},
  pages={481--490},
  year={2023},
  organization={Springer}
}

@inproceedings{wang2022mixed,
  title={Mixed transformer u-net for medical image segmentation},
  author={Wang, Hongyi and Xie, Shiao and Lin, Lanfen and Iwamoto, Yutaro and Han, Xian-Hua and Chen, Yen-Wei and Tong, Ruofeng},
  booktitle={ICASSP 2022-2022 IEEE international conference on acoustics, speech and signal processing (ICASSP)},
  pages={2390--2394},
  year={2022},
  organization={IEEE}
}

@inproceedings{xiao2018weighted,
  title={Weighted res-unet for high-quality retina vessel segmentation},
  author={Xiao, Xiao and Lian, Shen and Luo, Zhiming and Li, Shaozi},
  booktitle={2018 9th international conference on information technology in medicine and education (ITME)},
  pages={327--331},
  year={2018},
  organization={IEEE}
}

@inproceedings{zhou2018unet++,
  title={Unet++: A nested u-net architecture for medical image segmentation},
  author={Zhou, Zongwei and Rahman Siddiquee, Md Mahfuzur and Tajbakhsh, Nima and Liang, Jianming},
  booktitle={International workshop on deep learning in medical image analysis},
  pages={3--11},
  year={2018},
  organization={Springer}
}

@article{siegel2024cancer,
  title={Cancer statistics, 2024},
  author={Siegel, Rebecca L and Wagle, Neil C and Jemal, Ahmedin},
  journal={CA: a cancer journal for clinicians},
  volume={74},
  number={1},
  pages={12--49},
  year={2024},
  publisher={Wiley Periodicals, Inc. on behalf of the American Cancer Society}
}

@article{gershenwald2018melanoma,
  title={Melanoma staging: American joint committee on cancer (AJCC) and beyond},
  author={Gershenwald, Jeffrey E and Scolyer, Richard A},
  journal={Annals of surgical oncology},
  volume={25},
  number={8},
  pages={2105--2110},
  year={2018},
  publisher={Springer}
}

@article{esteva2017dermatologist,
  title={Dermatologist-level classification of skin cancer with deep neural networks},
  author={Esteva, Andre and Kuprel, Brett and Novoa, Roberto A and Ko, Justin and Swetter, Susan M and Blau, Helen M and Thrun, Sebastian},
  journal={nature},
  volume={542},
  number={7639},
  pages={115--118},
  year={2017},
  publisher={Nature Publishing Group UK London}
}

@article{mirikharaji2023survey,
  title={A survey on deep learning for skin lesion segmentation},
  author={Mirikharaji, Zahra and Abhishek, Kumar and Bissoto, Alceu and Barata, Catarina and Avila, Sandra and Valle, Eduardo and Celebi, M Emre and Hamarneh, Ghassan},
  journal={Medical Image Analysis},
  volume={88},
  pages={102863},
  year={2023},
  publisher={Elsevier}
}

@article{wu2024ultralight,
  title={Ultralight vm-unet: Parallel vision mamba significantly reduces parameters for skin lesion segmentation},
  author={Wu, Renkai and Liu, Yinghao and Ning, Guochen and Liang, Pengchen and Chang, Qing},
  journal={Patterns},
  year={2024},
  publisher={Elsevier}
}

@article{gu2023mamba,
  title={Mamba: Linear-time sequence modeling with selective state spaces},
  author={Gu, Albert and Dao, Tri},
  journal={arXiv preprint arXiv:2312.00752},
  year={2023}
}

@article{liao2024lightm,
  title={Lightm-unet: Mamba assists in lightweight unet for medical image segmentation},
  author={Liao, Weibin and Zhu, Yinghao and Wang, Xinyuan and Pan, Chengwei and Wang, Yasha and Ma, Liantao},
  journal={arXiv preprint arXiv:2403.05246},
  year={2024}
}

@inproceedings{peng2023rwkv,
  title={RWKV: Reinventing RNNs for the Transformer Era},
  author={Peng, Bo and Alcaide, Eric and Anthony, Quentin and Albalak, Alon and Arcadinho, Samuel and Biderman, Stella and Cao, Huanqi and Cheng, Xin and Chung, Michael and Derczynski, Leon and others},
  booktitle={Findings of the Association for Computational Linguistics: EMNLP 2023},
  pages={14048--14077},
  year={2023}
}

@misc{li2024ukanmakesstrongbackbone,
      title={U-KAN Makes Strong Backbone for Medical Image Segmentation and Generation}, 
      author={Chenxin Li and Xinyu Liu and Wuyang Li and Cheng Wang and Hengyu Liu and Yifan Liu and Zhen Chen and Yixuan Yuan},
      year={2024},
      eprint={2406.02918},
      archivePrefix={arXiv},
      primaryClass={eess.IV},
      url={https://arxiv.org/abs/2406.02918}, 
}

@article{zaki2024role,
  title={The role of point-of-care ultrasound (POCUS) imaging in clinical outcomes during cardiac arrest: a systematic review},
  author={Zaki, Hany A and Iftikhar, Haris and Shaban, Eman E and Najam, Mavia and Alkahlout, Baha Hamdi and Shallik, Nabil and Elnabawy, Wael and Basharat, Kaleem and Azad, Aftab Mohammad},
  journal={The Ultrasound Journal},
  volume={16},
  number={1},
  pages={4},
  year={2024},
  publisher={Springer}
}

@inproceedings{valanarasu2022unext,
  title={Unext: Mlp-based rapid medical image segmentation network},
  author={Valanarasu, Jeya Maria Jose and Patel, Vishal M},
  booktitle={International conference on medical image computing and computer-assisted intervention},
  pages={23--33},
  year={2022},
  organization={Springer}
}

@inproceedings{jiang2024lv,
  title={LV-UNet: A Lightweight and Vanilla Model for Medical Image Segmentation},
  author={Jiang, Juntao and Wang, Mengmeng and Tian, Huizhong and Cheng, Lingbo and Liu, Yong},
  booktitle={2024 IEEE International Conference on Bioinformatics and Biomedicine (BIBM)},
  pages={4240--4246},
  year={2024},
  organization={IEEE}
}

@article{loshchilov2017decoupled,
  title={Decoupled weight decay regularization},
  author={Loshchilov, I},
  journal={arXiv preprint arXiv:1711.05101},
  year={2017}
}

@article{loshchilov2016sgdr,
  title={Sgdr: Stochastic gradient descent with warm restarts},
  author={Loshchilov, Ilya and Hutter, Frank},
  journal={arXiv preprint arXiv:1608.03983},
  year={2016}
}

@inproceedings{liu2024rolling,
  title={Rolling-Unet: Revitalizing MLP’s Ability to Efficiently Extract Long-Distance Dependencies for Medical Image Segmentation},
  author={Liu, Yutong and Zhu, Haijiang and Liu, Mengting and Yu, Huaiyuan and Chen, Zihan and Gao, Jie},
  booktitle={Proceedings of the AAAI Conference on Artificial Intelligence},
  volume={38},
  pages={3819--3827},
  year={2024}
}

@article{ruan2024vm,
  title={Vm-unet: Vision mamba unet for medical image segmentation},
  author={Ruan, Jiacheng and Xiang, Suncheng},
  journal={arXiv preprint arXiv:2402.02491},
  year={2024}
}

@article{ma2024u,
  title={U-mamba: Enhancing long-range dependency for biomedical image segmentation},
  author={Ma, Jun and Li, Feifei and Wang, Bo},
  journal={arXiv preprint arXiv:2401.04722},
  year={2024}
}

@article{wang2024mamba,
  title={Mamba-unet: Unet-like pure visual mamba for medical image segmentation},
  author={Wang, Ziyang and Zheng, Jian-Qing and Zhang, Yichi and Cui, Ge and Li, Lei},
  journal={arXiv preprint arXiv:2402.05079},
  year={2024}
}

@inproceedings{liu2024swin,
  title={Swin-umamba: Mamba-based unet with imagenet-based pretraining},
  author={Liu, Jiarun and Yang, Hao and Zhou, Hong-Yu and Xi, Yan and Yu, Lequan and Li, Cheng and Liang, Yong and Shi, Guangming and Yu, Yizhou and Zhang, Shaoting and others},
  booktitle={International Conference on Medical Image Computing and Computer-Assisted Intervention},
  pages={615--625},
  year={2024},
  organization={Springer}
}

@inproceedings{codella2018skin,
  title={Skin lesion analysis toward melanoma detection: A challenge at the 2017 international symposium on biomedical imaging (isbi), hosted by the international skin imaging collaboration (isic)},
  author={Codella, Noel CF and others},
  booktitle={2018 IEEE 15th international symposium on biomedical imaging (ISBI 2018)},
  pages={168--172},
  year={2018},
  organization={IEEE}
}

@article{ma2024lmu,
  title={LMU-Net: lightweight U-shaped network for medical image segmentation},
  author={Ma, Ting and Wang, Ke and Hu, Feng},
  journal={Medical \& biological engineering \& computing},
  volume={62},
  number={1},
  pages={61--70},
  year={2024},
  publisher={Springer}
}

@article{li2024lite,
  title={Lite-UNet: A lightweight and efficient network for cell localization},
  author={Li, Bo and Zhang, Yong and Ren, Yunhan and Zhang, Chengyang and Yin, Baocai},
  journal={Engineering Applications of Artificial Intelligence},
  volume={129},
  pages={107634},
  year={2024},
  publisher={Elsevier}
}

@inproceedings{xu2024lb,
  title={Lb-unet: A lightweight boundary-assisted unet for skin lesion segmentation},
  author={Xu, Jiahao and Tong, Lyuyang},
  booktitle={International Conference on Medical Image Computing and Computer-Assisted Intervention},
  pages={361--371},
  year={2024},
  organization={Springer}
}

@inproceedings{dinh20231m,
  title={1M parameters are enough? A lightweight CNN-based model for medical image segmentation},
  author={Dinh, Binh-Duong and Nguyen, Thanh-Thu and Tran, Thi-Thao and Pham, Van-Truong},
  booktitle={2023 Asia Pacific Signal and Information Processing Association Annual Summit and Conference (APSIPA ASC)},
  pages={1279--1284},
  year={2023},
  organization={IEEE}
}

@inproceedings{rahman2025mk,
  title={MK-UNet: Multi-kernel Lightweight CNN for Medical Image Segmentation},
  author={Rahman, Md Mostafijur and Marculescu, Radu},
  booktitle={Proceedings of the IEEE/CVF International Conference on Computer Vision},
  pages={1042--1051},
  year={2025}
}

@misc{jiang2025rwkvunetimprovingunetlongrange,
      title={RWKV-UNet: Improving UNet with Long-Range Cooperation for Effective Medical Image Segmentation}, 
      author={Juntao Jiang and Jiangning Zhang and Weixuan Liu and Muxuan Gao and Xiaobin Hu and Zhucun Xue and Yong Liu and Shuicheng Yan},
      year={2025},
      eprint={2501.08458},
      archivePrefix={arXiv},
      primaryClass={eess.IV},
      url={https://arxiv.org/abs/2501.08458}, 
}

@inproceedings{chen2024tinyu,
  title={Tinyu-net: Lighter yet better u-net with cascaded multi-receptive fields},
  author={Chen, Junren and Chen, Rui and Wang, Wei and Cheng, Junlong and Zhang, Lei and Chen, Liangyin},
  booktitle={International Conference on Medical Image Computing and Computer-Assisted Intervention},
  pages={626--635},
  year={2024},
  organization={Springer}
}

@inproceedings{han2020ghostnet,
  title={Ghostnet: More features from cheap operations},
  author={Han, Kai and Wang, Yunhe and Tian, Qi and Guo, Jianyuan and Xu, Chunjing and Xu, Chang},
  booktitle={Proceedings of the IEEE/CVF conference on computer vision and pattern recognition},
  pages={1580--1589},
  year={2020}
}

@article{paszke2019pytorch,
  title={Pytorch: An imperative style, high-performance deep learning library},
  author={Paszke, Adam and Gross, Sam and Massa, Francisco and Lerer, Adam and Bradbury, James and Chanan, Gregory and Killeen, Trevor and Lin, Zeming and Gimelshein, Natalia and Antiga, Luca and others},
  journal={Advances in neural information processing systems},
  volume={32},
  year={2019}
}

@article{codella2019skin,
  title={Skin lesion analysis toward melanoma detection 2018: A challenge hosted by the international skin imaging collaboration (isic)},
  author={Codella, Noel and Rotemberg, Veronica and Tschandl, Philipp and Celebi, M Emre and Dusza, Stephen and Gutman, David and Helba, Brian and Kalloo, Aadi and Liopyris, Konstantinos and Marchetti, Michael and others},
  journal={arXiv preprint arXiv:1902.03368},
  year={2019}
}

@article{tschandl2018ham10000,
  title={The HAM10000 dataset, a large collection of multi-source dermatoscopic images of common pigmented skin lesions},
  author={Tschandl, Philipp and Rosendahl, Cliff and Kittler, Harald},
  journal={Scientific data},
  volume={5},
  number={1},
  pages={1--9},
  year={2018},
  publisher={Nature Publishing Group}
}

@article{mendoncca2015ph2,
  title={Ph2: A public database for the analysis of dermoscopic images},
  author={Mendon{\c{c}}a, Teresa and Celebi, M and Mendonca, T and Marques, J},
  journal={Dermoscopy image analysis},
  year={2015},
  publisher={CRC Press Boca Raton, FL, USA}
}

@inproceedings{nguyen2025mambau,
  title={MambaU-lite: A lightweight model based on Mamba and integrated channel-spatial attention for skin lesion segmentation},
  author={Nguyen, Thi-Nhu-Quynh and Ho, Quang-Huy and Nguyen, Duy-Thai and Le, Hoang-Minh-Quang and Pham, Van-Truong and Tran, Thi-Thao},
  booktitle={The International Conference on Intelligent Systems \& Networks},
  pages={49--58},
  year={2025},
  organization={Springer}
}

@inproceedings{tang2024cmunext,
  title={Cmunext: An efficient medical image segmentation network based on large kernel and skip fusion},
  author={Tang, Fenghe and Ding, Jianrui and Quan, Quan and Wang, Lingtao and Ning, Chunping and Zhou, S Kevin},
  booktitle={2024 IEEE International Symposium on Biomedical Imaging (ISBI)},
  pages={1--5},
  year={2024},
  organization={IEEE}
}

@inproceedings{zhao2022decoupled,
  title={Decoupled knowledge distillation},
  author={Zhao, Borui and Cui, Quan and Song, Renjie and Qiu, Yiyu and Liang, Jiajun},
  booktitle={Proceedings of the IEEE/CVF Conference on computer vision and pattern recognition},
  pages={11953--11962},
  year={2022}
}

@article{aldhabyani2020busi,
  title={Dataset of breast ultrasound images},
  author={Al-Dhabyani, Walid and Gomaa, Mohammed and Khaled, Hussien and Fahmy, Aly},
  journal={Data in Brief},
  volume={28},
  pages={104863},
  year={2020},
  publisher={Elsevier},
  doi={10.1016/j.dib.2019.104863}
}

\end{document}